\RequirePackage{fix-cm}
\documentclass[onecolumn,3p,times,numbers,sort]{elsarticle}
\usepackage{float}

\usepackage[utf8]{inputenc}  
\usepackage[T1]{fontenc}     

\usepackage{latexsym}
\usepackage{soul,color}
\usepackage{graphicx}
\usepackage{epstopdf}
\usepackage{multirow}
\usepackage{siunitx}
\usepackage{epsdice}
\usepackage{algorithm,algpseudocode}
\usepackage{hyperref}
\usepackage{url}
\usepackage{lipsum}
\usepackage[table]{xcolor}
\usepackage{colortbl}
\usepackage{amssymb}
\usepackage{svg}
\usepackage{booktabs}  
\usepackage{seqsplit}
\usepackage{paralist}
\usepackage{enumitem}
\usepackage{amsmath}
\usepackage{cleveref}
\usepackage{amsthm}
\usepackage{titlesec}

\titleformat{\paragraph}[runin]{\normalfont\bfseries}{\theparagraph}{1em}{}

\theoremstyle{plain}

\theoremstyle{definition}

\title{SechKAN: Kolmogorov–Arnold Networks with Hyperbolic Secant Functions}

\usepackage{xurl} 

\usepackage{pbox}

\setlist{noitemsep, topsep=0pt, partopsep=0pt, parsep=0pt, left=15pt}

\titlespacing*{\section}{0pt}{1.5ex plus 1ex minus .2ex}{1.5ex plus .2ex}
\titlespacing*{\subsection}{0pt}{1.5ex plus 1ex minus .2ex}{1.5ex plus .2ex}
\titlespacing*{\subsubsection}{0pt}{1.5ex plus 1ex minus .2ex}{1.5ex plus .2ex}

\begin{document}

\begin{frontmatter}
\author[aff1,aff2]{Hoang-Thang Ta}
\ead{thangth@uit.edu.vn}

\address[aff1]{University of Information Technology, Ho Chi Minh City, Vietnam}
\address[aff2]{Vietnam National University Ho Chi Minh City, Ho Chi Minh City, Vietnam}

\begin{abstract}

In recent years, Kolmogorov–Arnold Networks (KANs) have attracted increasing attention due to their effectiveness in machine learning and scientific computing, offering a new paradigm for neural network design. In this paper, we present SechKAN, a novel KAN based on hyperbolic secant (sech) functions. The hyperbolic secant basis is adopted for its smooth bell-shaped form, localized responses, and well-behaved gradients. We employ a 1D linear projection to reduce the number of parameters, allowing SechKAN to maintain a model size comparable to that of multilayer perceptrons (MLPs). Experimental results show the effectiveness of SechKAN on function fitting, PDE surrogate modeling, and image classification benchmarks, including MNIST, Fashion-MNIST, CIFAR-10, and CIFAR-100. On function fitting, SechKAN achieves performance comparable to both MLPs and representative KAN variants. On PDE surrogate modeling, it outperforms MLPs and achieves competitive or better performance than representative KAN variants. On image classification benchmarks, SechKAN achieves the best performance among the evaluated KAN variants while remaining competitive with MLPs using a comparable number of parameters. However, SechKAN still incurs higher computational cost than MLPs and some KAN variants. Our source code is publicly available at \url{https://github.com/hoangthangta/All-KAN}.

\textcolor{blue}{Please refer to the journal version at \url{https://www.sciencedirect.com/science/article/pii/S0925231226024665} for the most accurate and up-to-date version.}

\end{abstract}


\end{frontmatter}

\section{Introduction}

For decades, multilayer perceptrons (MLPs) have been widely used in neural network models to learn features from data. They can be applied as standalone models or combined with other architectures such as CNNs, RNNs, and LSTMs due to their simplicity and speed. However, MLPs are not always the most suitable choice for modeling complex nonlinear relationships or smooth functional patterns in data, which may require alternative model structures. Kolmogorov–Arnold Networks (KANs) have recently attracted considerable attention as a promising neural network architecture. With their design using polynomial basis functions, they can capture many types of data effectively and perform well on various mathematics, AI, and physics tasks.

Although KANs have proven effective across many tasks and applications, they still have several limitations, including the significantly larger number of parameters in their layers than in MLPs, which results in longer training times. Furthermore, even with the same number of parameters, their performance is sometimes not better than MLPs in certain tasks. Some basis functions used in KANs, such as the B-spline basis adopted in EfficientKAN~\cite{Blealtan2024}, involve recursive and piecewise computations, which may increase implementation complexity on GPUs. Our goal is to explore alternative basis functions with simpler formulations that may reduce computational overhead and simplify implementation compared with those of recursive spline-based basis functions. Also, we aim to develop a neural network that maintains approximately the same number of parameters as MLPs while achieving competitive or improved performance.

In this paper, we introduce SechKAN, a novel KAN that adopts the hyperbolic secant (sech) function as the basis function. The hyperbolic secant provides a smooth, localized, and infinitely differentiable basis with balanced decay, which may improve training stability. Similar to other KAN variants, this basis expansion does not by itself reduce the number of parameters. Instead, SechKAN employs a lightweight 1D linear projection as a structural bottleneck to aggregate the basis responses, thereby reducing model complexity while preserving the expressive power of KANs. Recent studies have highlighted the sensitivity of Gaussian-based KANs to basis scale selection \cite{noorizadegan2026making,noorizadegan2026partition}; accordingly, our comparisons use the original or default configurations of FastKAN, FasterKAN, and other representative KAN variants, with only minor modifications for a unified experimental protocol.

The effectiveness of SechKAN is evaluated on three tasks: function fitting, PDE surrogate modeling, and image classification. Our main contributions are as follows:
\begin{itemize}
    \item We propose SechKAN, a novel KAN that combines hyperbolic secant (sech) basis functions with a lightweight 1D linear projection, enabling parameter-efficient nonlinear representation learning while maintaining a model size comparable to that of MLPs.

    \item We derive the parameter count and FLOPs of SechKAN and evaluate its computational efficiency, including GPU memory consumption, forward and backward execution time, training throughput, and gradient flow.

   \item We show that SechKAN achieves competitive or better performance across function fitting, PDE surrogate modeling, and image classification. In particular, it consistently outperforms representative KAN variants on image classification while remaining competitive with MLPs.

    \item We conduct extensive ablation studies on grid size, activation functions, and normalization strategies, providing practical guidelines for configuring SechKAN across different tasks.

\end{itemize}

Aside from this introduction, the remainder of the paper is organized as follows. Section 2 reviews related work on KANs and their limitations in terms of parameter efficiency. Section 3 introduces KANs and several representative variants. Section 4 presents SechKAN, including its design and parameterization, a comparison of the hyperbolic secant basis function with existing KAN bases, and the motivation for parameter reduction. It also compares the parameter count and FLOPs of SechKAN with those of existing KAN variants and MLPs. Section 5 describes the evaluation metrics, computational complexity, and experimental setup, and presents the results for function fitting, PDE surrogate modeling, and image classification tasks. Section 6 presents ablation studies on the individual contributions of the sech basis function and the 1D projection in SechKAN, followed by analyses of grid sizes, activation functions, and data normalization strategies. Section 7 discusses the limitations of the proposed approach and provides practical guidelines for selecting the grid sizes and data normalization types. Finally, Section 8 concludes the paper and outlines directions for future research. The appendices provide a comparison of the hyperbolic secant function with other activation functions used in neural networks, detailed experimental settings for function fitting and PDE surrogate modeling, additional image classification experiments, and an analysis of gradient flow and saturation in SechKAN.

\section{Related Work}\label{sec_related_works}

KANs were recently introduced by \citet{liu2025kan}, providing a new perspective on neural network design by replacing fixed activation functions with learnable univariate functions parameterized using B-splines. KANs are based on the Kolmogorov-Arnold representation theorem (KART), which is closely related to the solution of Hilbert's 13th problem and states that any continuous multivariate function can be represented as a finite composition and summation of one-dimensional functions~\cite{kolmogorov1957representation,braun2009constructive}. Prior to KAN, KART had already inspired the development of several neural network architectures, including spline-based  models~\cite{leni2013kolmogorov,van2022kasam,igelnik2003kolmogorov,fakhoury2022exsplinet}.

KANs have been applied across several domains, including differential equation solving~\cite{wang2025kolmogorov,koenig2024kan}, time series forecasting~\cite{genet2024temporal,xu2024kolmogorov,vaca2024kolmogorov}, computer vision~\cite{li2025u,cheon2024demonstrating,ge2025tc}, quantum computing~\cite{kundu2024kanqas}, and mechanics simulations~\cite{abueidda2025deepokan}. Another active research direction explores replacing the original B-spline basis with alternative functions to improve expressiveness and adaptability~\cite{somvanshi2025survey}. Proposed alternatives include modified B-splines~\cite{Blealtan2024,ta2024bsrbf}, polynomial families such as Chebyshev and Legendre polynomials~\cite{torchkan,ss2024chebyshev} and other polynomial variants~\cite{teymoor2024exploring}, radial basis functions~\cite{li2024kolmogorov,ta2024bsrbf,abueidda2025deepokan}, activation functions~\cite{chen2024lss, qiu2025relu}, and Fourier bases~\cite{xu2024fourierkan}, wavelets~\cite{bozorgasl2024wav,seydi2024unveiling}, rational and fractional Jacobi functions~\cite{aghaei2026rkan,aghaei2025fkan}, and customized activation functions~\cite{qiu2025relu,athanasios2024}.

Several studies have addressed the parameter inefficiency of KANs, motivating the development of lightweight architectures with fewer parameters. For example, \citet{yang2025kolmogorov} introduced Group KAN in Kolmogorov-Arnold Transformers (KATs), which reduces the number of parameters and computational cost through weight sharing among groups of edges. Similarly, \citet{ta2025prkan} proposed PRKAN, which employs multiple parameter-reduction techniques to achieve model sizes comparable to those of MLPs. More recently, several parameter-efficient KANs have been proposed. GS-KAN reduces parameter complexity by generating edge functions from a shared learnable parent function through linear transformations~\cite{eliasson2025gs}. LeanKAN introduces a compact replacement for AddKAN and MultKAN with fewer parameters and hyperparameters, improving memory efficiency and learning capability~\cite{koenig2025leankan}. GroupKAN improves parameter efficiency through group-structured spline modeling and shared spline functions, reducing parameter redundancy while preserving the expressive capacity of KANs~\cite{li2025groupkan}. 

In KANs, B-spline basis functions implemented via the de Boor--Cox recursion require substantially more computations than simpler basis functions, often leading to higher computational cost than MLPs~\cite{qiu2025powermlp,li2024kolmogorov}. To address this limitation, many studies have explored computationally efficient alternatives, including radial basis functions (RBFs)~\cite{li2024kolmogorov} and activation functions~\cite{chen2024lss}. Despite these advances, achieving both parameter efficiency and computational efficiency remains an open challenge for KANs. This motivates the exploration of alternative basis functions and parameterization strategies that can reduce model complexity while maintaining strong predictive performance.

\section{Background}
\label{sec:background}

\subsection{Kolmogorov-Arnold Network}
\label{KAN}

Kolmogorov–Arnold Networks (KANs) are grounded in the Kolmogorov–Arnold Representation Theorem (KART), which ensures that any continuous multivariate function over a bounded domain can be expressed as a finite sum of single-variable functions. Let $\mathbf{x} = (x_1, \ldots, x_n) \in [0,1]^n$. Any continuous function $f : [0,1]^n \to \mathbb{R}$ can be represented as~\cite{ismailov2025addressing}:

\begin{equation}
\begin{aligned}
f(\mathbf{x}) = f(x_1, \ldots, x_n) = \sum_{q=1}^{2n+1} \Phi_q \left( \sum_{p=1}^{n} \phi_{q,p}(x_p) \right)
\end{aligned}
\label{eq:kart}
\end{equation}
Here, each outer function \(\Phi_q\) aggregates an inner summation of continuous transformations \(\phi_{q,p}(x_p)\). This two-level decomposition, in which outer functions combine inner single-variable mappings, forms the theoretical basis that allows KANs to approximate arbitrary multivariate functions using only univariate components and summations.

\citet{liu2025kan} introduced the Kolmogorov–Arnold Network (KAN), extending the Kolmogorov–Arnold theorem’s two-layer structure with \(2n+1\) hidden terms to arbitrary widths and depths, thereby enhancing expressive power for machine learning. This relies on the choice of functions \(\Phi_q\) and \(\phi_{q,p}\), as defined in \Cref{eq:kart}. A KAN with \(L\) layers applies successive transformations through function matrices \(\Phi_0, \Phi_1, \dots, \Phi_{L-1}\), yielding:

\begin{equation}
\begin{aligned}
\text{KAN}(\mathbf{x}) = (\Phi_{L-1} \circ \Phi_{L-2} \circ \cdots \circ \Phi_1 \circ \Phi_0)\mathbf{x}
\end{aligned}
\label{eq:kan}
\end{equation}

Each function matrix \(\Phi_l\) consists of pre-activations defined by activation functions \(\phi_{l,j,i}\), which connect neuron \(\,(l,i)\) to neuron \(\,(l+1,j)\):  

\begin{equation}
\begin{aligned}
\phi_{l,j,i}, \quad l = 0, \cdots, L - 1, \quad i = 1, \cdots, n_l, \quad j = 1, \cdots, n_{l+1}
\end{aligned}
\label{eq:acti_funct}
\end{equation}

With \(n_l\) nodes in the \(l^{th}\) layer, the transformation from \(\mathbf{x}_l\) to \(\mathbf{x}_{l+1}\) is computed by the function matrix \(\Phi_l \in \mathbb{R}^{n_{l+1}\times n_l}\):  

\begin{equation}
\begin{aligned}
\mathbf{x}_{l+1} = 
\underbrace{\left(
\begin{array}{cccc}
\phi_{l,1,1}(\cdot) & \phi_{l,1,2}(\cdot) & \cdots & \phi_{l,1,n_l}(\cdot) \\
\phi_{l,2,1}(\cdot) & \phi_{l,2,2}(\cdot) & \cdots & \phi_{l,2,n_l}(\cdot) \\
\vdots & \vdots & \ddots & \vdots \\
\phi_{l,n_{l+1},1}(\cdot) & \phi_{l,n_{l+1},2}(\cdot) & \cdots & \phi_{l,n_{l+1},n_l}(\cdot)
\end{array}\right)}_{\Phi_{l}} \mathbf{x}_l
\label{eq:function_matrix}
\end{aligned}
\end{equation}


\citet{liu2025kan} constructed KAN using a residual activation function \(\phi(x)\), defined as the weighted sum of a base function and a spline function, with weight matrices \(w_b\) and \(w_s\), respectively:

\begin{equation}
\begin{aligned}
\phi(x) = w_b b(x) + w_s spline(x)
\end{aligned}
\label{eq:acti_funct_imp}
\end{equation}

\begin{equation}
\begin{aligned}
b(x) = silu(x) = \frac{x}{1 + e^{-x}}
\end{aligned}
\label{eq:b_function}
\end{equation}

\begin{equation}
\begin{aligned}
spline(x) = \sum_{i}c_iB_i(x)
\end{aligned}
\label{eq:spline_function}
\end{equation}

In \Cref{eq:acti_funct_imp}, \(b(x)\) is the SiLU function (\Cref{eq:b_function}), and \(spline(x)\) is a linear combination of B-splines with coefficients \(c_i\) (\Cref{eq:spline_function}). The activation is initialized with \(w_s = 1\) (so \(spline(x) \approx 0\)) and \(w_b\) using initializations such as Xavier.

\begin{figure*}[htbp]
  \centering
\includegraphics[scale=0.8]{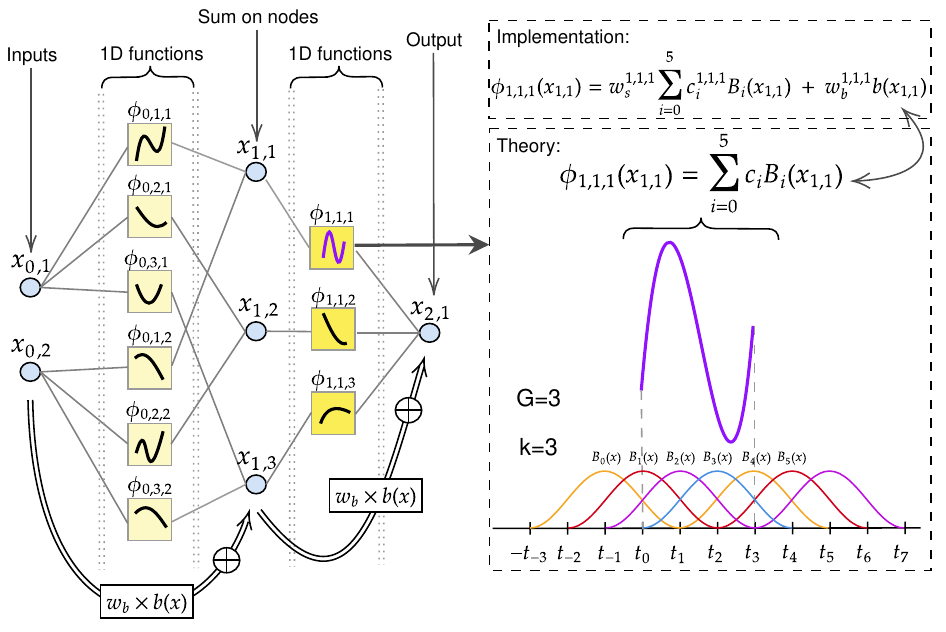}
  \centering
  \caption{Left: The structure of KAN(2,3,1). Right: Calculate \(\phi_{1,1,1}\) by using control points and B-splines. \(G \) and \( k \) represent the grid size and the spline order, while the number of B-splines, \(n\), is given by \( G + k = 3 + 3 = 6\). The function $\phi$ is calculated differently in theory and in implementation, where it is multiplied by its weight matrix and then added to a linear projection. The figure was modified from~\citet{ta2026fc}.}
\label{fig:kan_diagram}
\end{figure*}

\Cref{fig:kan_diagram} shows the architecture of KAN(2,3,1) with 2 input nodes, 3 hidden nodes, and 1 output node. Each node computes its output as the sum of functions $\phi$ along its incoming edges. The figure also illustrates how $\phi$ is constructed from control points and B-splines, where the number of B-splines is given by $G + k = 3 + 3 = 6$, so the index $i$ ranges from 0 to 5. The theoretical formula of $\phi$ differs from its implementation in KAN (\Cref{eq:acti_funct_imp}), which involves multiplication with its spline matrix $w_s$, the base weight matrix $w_b$, and the activation function $b(x)$ applied to the input $x$.

\subsection{KAN variants}
\begin{table*}[htbp]
\centering
\caption{Summary of KAN variants, their defining base and inner functions.}
\label{tab:kan_variants}
\begin{tabular}{lp{5cm}p{7cm}}
\hline
\textbf{Variant} & \textbf{Inner Function} & \textbf{Base Function} \\ \hline

EfficientKAN~\cite{Blealtan2024} & 
$\phi(x) = w_b b(x) + w_s B(x)$ & 
$\mathit{B}(x) = \sum_{i}{c_i B_i(x)}, \quad c_i =$ spline coefficient \\ 
& & 
$B_i(x) \equiv B_{i,k}(x) \text{ for chosen order } k$ \\ 
& & 
$B_{i,0}(x)  =
\begin{cases}
1, & t_i \leq x < t_{i+1} \\
0, & \text{otherwise}
\end{cases}$ \\ 
& & 
$B_{i,k}(x) = \frac{x - t_i}{t_{i+k} - t_i} B_{i,k-1}(x) 
+ \frac{t_{i+k+1} - x}{t_{i+k+1} - t_{i+1}} B_{i+1,k-1}(x)$ \\ \hline

FastKAN~\cite{li2024kolmogorov} & 
$\phi(x) = w_b b(x) + w_s \mathit{RBF}(x)$ & 
$\mathit{RBF}(x) = \exp\!\left(-\tfrac{\|x - d\|^2}{2h^2}\right),$ \\ 
 & &
$d =$ center, $h =$ function width or spread \\ \hline

FasterKAN~\cite{athanasios2024} & 
$\phi(x) = w_b b(x) + w_s \mathit{RSWAF}(x)$ & 
$\mathit{RSWAF}(x) = 1 - \left(\tanh\!\left(\tfrac{\|x - d\|}{h}\right)\right)^2,$ \\ 
 & &
$d =$ center, $h =$  function width or spread \\ \hline

BSRBF-KAN~\cite{ta2024bsrbf} & 
$\phi(x) = w_b b(x) + w_s \big(\mathit{B}(x) + \mathit{RBF}(x)\big)$ & 
$\mathit{B}(x)$ and $\mathit{RBF}(x)$ as defined above, parameters $c_i$, $d$, $h$ ~retain same roles in EfficientKAN and FastKAN \\ \hline

ReLU-KAN~\cite{qiu2025relu} & 
$\phi(x) = \mathcal{K} \ast R  = \sum_i w_i R_i(x),$ 
& 
$R_i(x) = \big[\operatorname{ReLU}(x - l_i) \cdot \operatorname{ReLU}(h_i - x)\big]^2 
\cdot \tfrac{16}{(h_i - l_i)^4},$ \\
& $\mathcal{K} = $ convolution kernel, \quad $w_i =$ learnable coefficients from the convolution layer
& where $l_i, h_i =$ learnable phase low/high boundaries \\
\hline
\end{tabular}
\end{table*}
KANs have inspired several variants designed to improve efficiency, scalability, and adaptability. In this section, we focus on those most relevant to our work and used for comparison in the experiments. EfficientKAN~\cite{Blealtan2024} refines the original design by employing linear combinations of B-splines with learnable scaling factors and replacing input regularization with weight regularization, thereby reducing both memory and computational overhead. FastKAN~\cite{li2024kolmogorov} builds on this by adding layer normalization and replacing B-splines with Gaussian radial basis functions (Gaussian RBFs), which accelerates training. FasterKAN~\cite{athanasios2024} further improves efficiency by introducing Reflectional Switch Activation Functions (RSWAF) that preserve accuracy while lowering the cost of forward and backward passes. To merge the benefits of splines and Gaussian RBFs, BSRBF-KAN~\cite{ta2024bsrbf} integrates both within each layer, achieving faster convergence, though sometimes at the risk of overfitting. Finally, ReLU-KAN~\cite{qiu2025relu} departs from spline- and RBF-based constructions by introducing bell-shaped basis functions derived from ReLU activations with learnable phase boundaries, enhancing both flexibility and generalization. A summary of their defining inner and base functions is provided in \Cref{tab:kan_variants}.

\section{SechKAN}

\subsection{Architecture}

\begin{figure*}[htbp]
\centering
\begin{minipage}[t]{0.61\textwidth}
    \vspace{0pt}
    \raggedright
    \includegraphics[width=\linewidth]{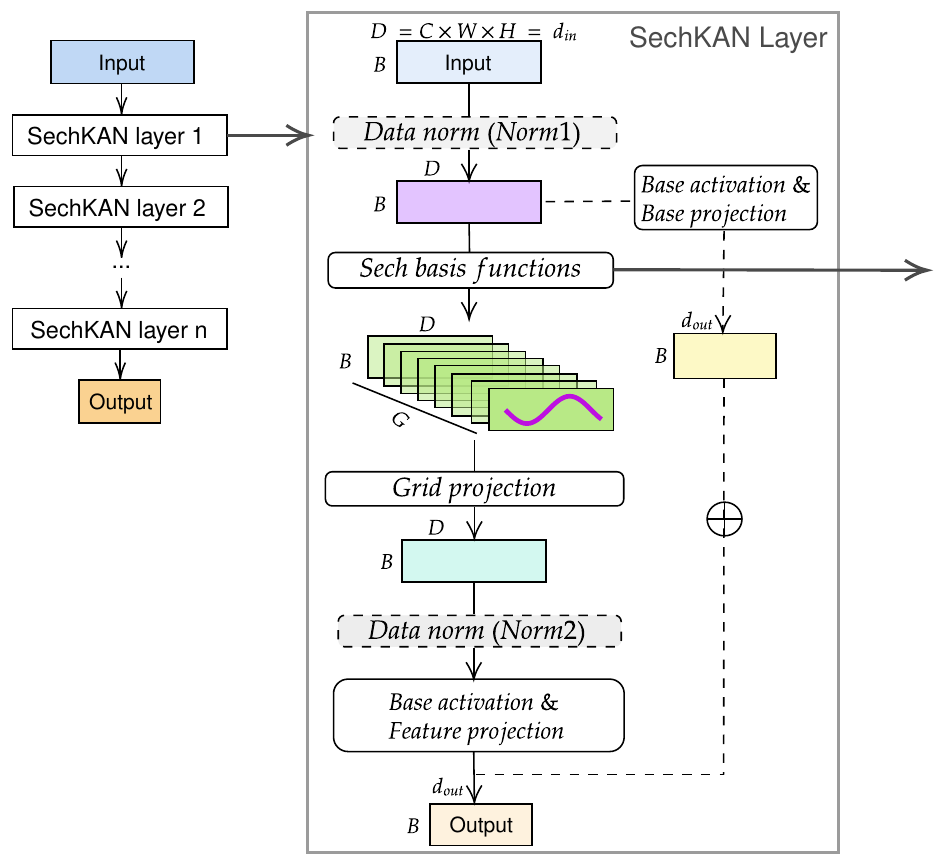}
\end{minipage}
\hfill
\begin{minipage}[t]{0.38\textwidth}
    \vspace{0pt}
    \raggedleft
    \includegraphics[width=\linewidth]{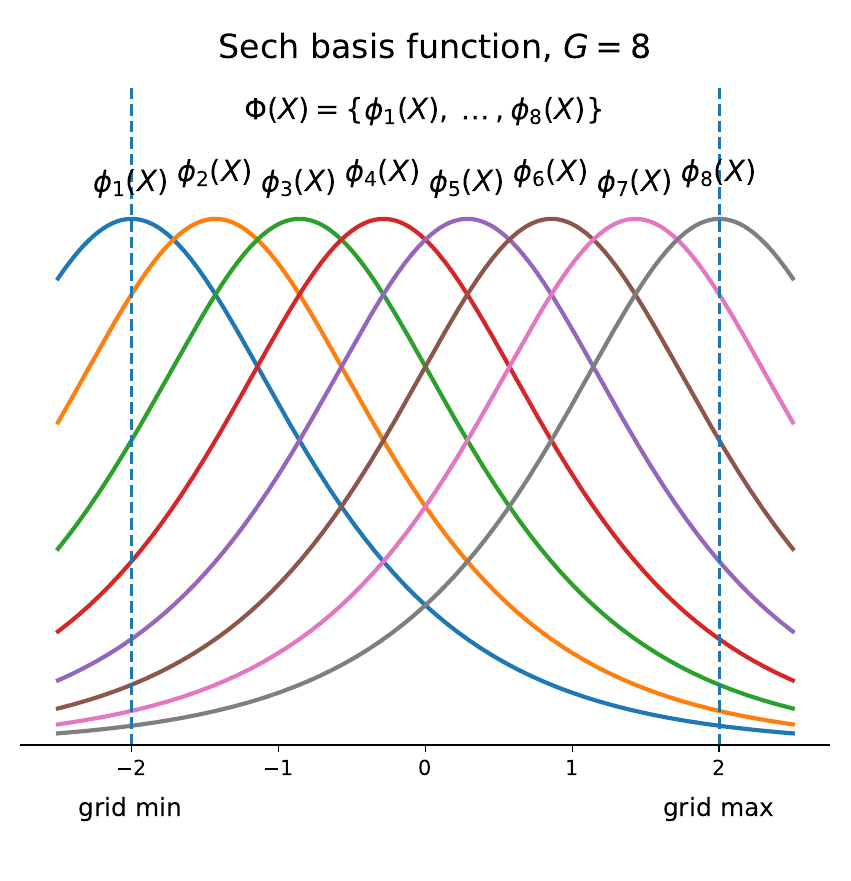}
\end{minipage}

\caption{SechKAN with multiple layers, and the architecture of a single SechKAN layer using sech basis functions.}
\label{fig:sechkan_archi}
\end{figure*}

As shown in \Cref{fig:sechkan_archi}, SechKAN, like other neural networks, consists of multiple layers. Each SechKAN layer takes an input and produces an output, which then serves as the input to the next layer, following the standard composition of neural network architectures. The input may optionally pass through a data normalization step (Norm1), followed by sech basis functions to obtain the standard KAN representation with an additional grid dimension $G$. The sech basis $\Phi$ consists of $G$ functions ${\phi_1, \phi_2, \dots, \phi_G}$ positioned within the range defined by the grid minimum and grid maximum.

A grid projection is then applied to reduce the last dimension from $G$ to 1, thereby lowering the number of parameters. As a result, the number of parameters in each layer remains comparable to that of MLPs. The transformed input is then passed through another optional normalization step (Norm2), followed by an activation function (e.g., SiLU) and a feature projection to produce the output. In parallel, a skip projection processes the normalized input to generate an additional output, which can be optionally added to the output of the feature projection.

The sech function is smooth, bell-shaped, exponentially decaying, and infinitely differentiable, promoting stable gradient behavior. It yields bounded outputs in $(0,1]$, ensuring stable activations and localized representations, with implicit regularization but potential vanishing gradients and loss of sign information. A detailed empirical analysis of gradient propagation, saturation behavior, and the effects of different normalization strategies is provided in \Cref{sec:sech_kan_gradient_flow}. Therefore, it is sufficient to control the input range before applying the sech functions. These properties support efficient and stable grid aggregation in SechKAN. Sech basis functions share similar characteristics with RBFs and B-splines used in KANs, while offering improved runtime efficiency, as analyzed in \Cref{sec:sech_analysis}.

\subsection{SechKAN Layer Formulation}
In this work, SechKAN is designed for one-dimensional input data. For two-dimensional or higher-dimensional inputs, the data are reshaped into one-dimensional vectors before being processed by the model.

Let $X \in \mathbb{R}^{B \times d_{in}}$ denote the input, where $B$ is the batch size and $d_{in}$ is the input dimension. SechKAN consists of multiple stacked layers, each following the architecture illustrated in \Cref{fig:sechkan_archi}. Passing $X$ through a SechKAN layer yields an output $Y \in \mathbb{R}^{B \times d_{\mathrm{out}}}$, where $d_{\mathrm{out}}$ is the output dimension, $g_{\min}$ and $g_{\max}$ denote the fixed minimum and maximum grid values, respectively, and $G$ denotes the grid size (i.e., the number of grid points).

First, $X$ is normalized using a normalization method (e.g., LayerNorm, BatchNorm, RMSNorm, or Min--Max normalization) to obtain $X'$, denoted as Norm1:
\begin{equation}
X' = \mathrm{Norm}_1(X)
\end{equation}
When Min--Max normalization is employed, $X$ is rescaled to the interval $[g_{\min}, g_{\max}]$ as follows:
\begin{equation}
X' =
\frac{X - \min(X)}
{\max(X) - \min(X) + \epsilon}
(g_{\max} - g_{\min})
+ g_{\min},
\end{equation}
where $\epsilon$ is a small positive constant introduced for numerical stability.


The normalized input $X'$ is then passed to the hyperbolic secant basis expansion. We distinguish between the grid resolution and the grid-center locations. The grid resolution $G$, which specifies the number of grid centers, is treated as a fixed architectural hyperparameter. For a given $G$, the learnable grid-center locations are represented by the vector
\begin{equation}
\mathcal{C} = \{c_g\}_{g=1}^{G},
\end{equation}
which are initialized uniformly according to
\begin{equation}
c_g = g_{\min} + \frac{g - 1}{G - 1}(g_{\max} - g_{\min}),
\quad g = 1, \dots, G,
\end{equation}
yielding $G$ uniformly spaced initial grid centers over the interval $[g_{\min}, g_{\max}]$. During training, the grid-center locations $\mathcal{C}$ are optimized jointly with the remaining model parameters, while the grid resolution $G$ remains fixed.

Two learnable scalars, a scale $s$ and a bias $\delta$, are applied to each basis response. The sech basis $\Phi = {\phi_1, \phi_2, \dots, \phi_G}$ is computed as
\begin{equation}
\Phi(X') = s \cdot
\frac{1}{\cosh\!\left(\frac{X' - \mathcal{C}}{w}\right)} + \delta,
\quad \Phi(X') \in \mathbb{R}^{B \times d_{in} \times G}
\label{eq:sech_basis}
\end{equation}
where $w$ is a single learnable scalar width parameter shared by all sech basis functions in a layer. By default, it is initialized to 1.0. To ensure that the width parameter remains positive, we optimize an unconstrained parameter $\theta_w$ instead of $w$ directly, with the width computed as $w=\exp(\theta_w)+10^{-8}$, where $10^{-8}$ prevents division by zero.

Subsequently, a grid-wise linear projection aggregates the $G$ basis responses of each feature:
\begin{equation}
Z = \Phi(X') W_{G} + b_{G}, \quad Z \in \mathbb{R}^{B \times d_{in}}
\end{equation}
where $W_{G} \in \mathbb{R}^{G \times 1}$ and $b_{G} \in \mathbb{R}$ are learnable parameters. This operation compresses the $G$ basis responses into a single scalar ($G \rightarrow 1$) through a lightweight 1D linear projection, enabling a more parameter-efficient layer design. While various parameter-reduction techniques are available, we adopt this approach because of its simplicity and computational efficiency.

Next, a second normalization step (Norm2) is applied:
\begin{equation}
Z' = \mathrm{Norm}_2(Z)
\end{equation}
An element-wise nonlinear activation function $\sigma(\cdot)$ is then applied, followed by a feature-wise linear projection:
\begin{equation}
Y = \sigma(Z') W_{F} + b_{F}, \quad Y \in \mathbb{R}^{B \times d_{out}}
\end{equation}
where $W_{F} \in \mathbb{R}^{d_{in} \times d_{out}}$ and $b_{F} \in \mathbb{R}^{d_{out}}$ are learnable parameters.

In parallel, a skip projection branch operates on the normalized input $X'$, applying the same activation function followed by a base linear projection. The resulting output is added to $Y$. If this branch is not used, the layer directly returns $Y$.

\subsection{Comparison of Hyperbolic Secant Basis Functions with Existing KAN Bases}

\begin{table*}[!ht]
\centering
\caption{Comparison of SechKAN and KAN variants based on their basis characteristics and parameter-reduction strategies. '--' indicates that no parameter reduction method is used.}
\label{tab:basis_comparison}
\begin{tabular}{
>{\raggedright\arraybackslash}p{4.8cm}
>{\raggedright\arraybackslash}p{5.0cm}
>{\raggedright\arraybackslash}p{2.8cm}
}
\hline
\textbf{Basis Function (KAN Variant)} &
\textbf{Basis Characteristics} &
\textbf{Parameter Reduction} \\
\hline

B-spline (EfficientKAN~\cite{Blealtan2024}) &
Piecewise polynomial, compact support, piecewise smooth, efficient B-spline implementation &
-- \\
\hline

Gaussian RBF (FastKAN~\cite{li2024kolmogorov}) &
Infinitely smooth, Gaussian decay, smooth bounded gradients, analytic closed form &
-- \\
\hline

RSWAF (FasterKAN~\cite{athanasios2024}) &
Infinitely smooth, tanh-based localization, smooth bounded gradients, analytic closed form &
-- \\
\hline

ReLU-derived basis (ReLU-KAN~\cite{qiu2025relu}) &
Piecewise quartic polynomial, compact support, piecewise smooth, analytic closed form &
-- \\
\hline

Hyperbolic secant or sech (SechKAN, Ours) &
Infinitely smooth, exponential decay, smooth bounded gradients, analytic closed form &
1D linear projection \\
\hline

B-spline and Gaussian RBF (PRKAN~\cite{ta2025prkan}) &
Original B-spline and Gaussian RBF basis with dimensionality reduction &
Attention, Conv, Conv+Pool, Dim-Sum, Feature Weight Vector \\
\hline

B-spline (LeanKAN~\cite{koenig2025leankan}) &
B-spline basis with multiplicative interactions &
Multiplicative operators \\
\hline

\end{tabular}
\end{table*}

Several KAN variants retain the overall KAN framework while replacing the original B-spline basis with alternative localized edge basis functions. Representative examples include EfficientKAN (an efficient implementation of B-splines), FastKAN (Gaussian RBFs), FasterKAN (RSWAFs), and ReLU-KAN (ReLU-derived localized basis functions with convolution-based aggregation). In addition to basis function design, recent studies have explored architectural modifications to improve efficiency. For example, PRKAN reduces the dimensionality of the basis representation using multiple strategies such as attention mechanisms, convolutional layers, convolution with pooling, dimension summation, and feature weight vectors, while LeanKAN incorporates multiplicative operations to reduce the parameter overhead.

SechKAN distinguishes itself by introducing the hyperbolic secant ($  \textit{sech}(x)  $) as a novel analytical basis function. Compared to Gaussian RBFs, $ \textit{sech}(x) $ exhibits smooth exponential decay and provides smooth local gradients within the normalized input range. Both the function and its derivative, $-\textit{sech}(x)\textit{tanh}(x)$, are smooth and decay exponentially. A more comprehensive empirical study of gradient magnitudes, saturation, and normalization effects is presented in \Cref{sec:sech_kan_gradient_flow}. In addition, SechKAN employs a lightweight 1D linear projection for basis dimensionality reduction, similar to the Feature Weight Vector strategy in PRKAN~\cite{ta2025prkan}. Consequently, SechKAN integrates a new basis function with a simple yet effective parameter-efficient design within the KAN framework. These distinctions in basis properties (decay rate, localization, gradient behavior) and architectural efficiency are summarized in Table~\ref{tab:basis_comparison}.

\subsection{Motivation for Parameter Reduction}

In traditional KANs, the basis functions expand each input feature into an intermediate representation of size $(B, d_{in}, G+k)$, where $G$ is the grid size and $k$ is the order of the basis functions. This representation is then reshaped to $(B, d_{in}(G+k))$ and projected by a weight matrix of size $(d_{in}(G+k), d_{out})$ to produce the output of size $(B, d_{out})$. Therefore, the number of learnable parameters increases with the number of basis functions $(G+k)$, resulting in high parameter complexity.

Motivated by the observation that the basis expansion already provides a nonlinear representation for each input feature, SechKAN introduces a learnable aggregation mechanism within each feature's basis space. The $G$ basis responses are compressed through a learnable one-dimensional projection, reducing the representation from $(B, d_{in}, G)$ to $(B, d_{in}, 1)$, which is then squeezed to $(B, d_{in})$. The resulting compact representation is then projected using a smaller weight matrix of size $(d_{in}, d_{out})$, substantially reducing the number of learnable parameters. This parameter-reduction strategy is inspired by PRKAN~\cite{ta2025prkan}.

From this perspective, SechKAN decomposes the output mapping into two stages: (i) learnable projection of the basis responses within each input feature and (ii) linear mixing across input dimensions. Thus, unlike traditional KANs, where each input-output edge has an independently parameterized nonlinear function, SechKAN first constructs a shared learnable univariate basis representation for each input feature and then combines these representations through linear feature mixing. Therefore, SechKAN can be viewed as a parameter-efficient and constrained variant of the KAN formulation that preserves its additive univariate functional structure while sharing part of the edge-function parameterization to reduce the number of parameters.

\subsection{Number of Parameters for SechKAN, KAN, and MLP}


Let $d_{\text{in}}$ and $d_{\text{out}}$ denote the input and output dimensions of a layer, respectively. Let $G$ denote the grid size (i.e., the number of grid points), and $K$ denote the spline order.

The number of parameters of an MLP layer is~\cite{yu2024kan}:
\begin{equation}
\begin{aligned}
\text{MLP}_{\text{param}} &= d_\text{in} d_\text{out} + d_\text{out}
\end{aligned}
\label{eq:mlp_params}
\end{equation}
The number of parameters of a KAN layer, counting the weight matrix including B-spline control points, shortcut weights, B-spline weights, plus bias, is~\cite{yu2024kan}:
\begin{equation}
\begin{aligned}
\text{KAN}_{\text{param}} &= d_\text{in} d_\text{out} (G + K + 3) + d_\text{out}
\end{aligned}
\label{eq:kan_params_full}
\end{equation}
If counting only the weight matrix and the bias term, it contains 
$d_{\mathrm{in}}d_{\mathrm{out}}(G+K)+d_{\mathrm{out}}$ 
trainable parameters~\cite{ta2025prkan}.


In SechKAN, the hyperbolic secant basis parameters are shared over the entire layer. Each layer contains a single learnable grid vector $G$, one scalar global scale parameter $s$, one global scalar bias parameter $\delta$, and, when enabled, one scale global width parameter $w$. These parameters are shared by all input features and output neurons; they are not feature-specific or edge-specific.

The sech basis therefore contributes
\begin{equation}
\mathrm{SechBasis}_{\mathrm{param}}
=
G+2+\mathbb{I}_{\mathrm{width}}
\end{equation}
trainable parameters, where $\mathbb{I}_{\mathrm{width}}\in\{0,1\}$ indicates whether the learnable width parameter $w$ is enabled.

The grid projection linearly combines the $G$ basis responses into a single scalar output. This projection consists of $G$ learnable weights and one bias parameter, resulting in:
\begin{equation}
\mathrm{Grid}_{\mathrm{param}}
=
G+1
\end{equation}
parameters. The feature projection layer has the same parameterization as an MLP layer:
\begin{equation}
\mathrm{Feature}_{\mathrm{param}}
=
d_{\mathrm{in}}d_{\mathrm{out}}
+d_{\mathrm{out}}
\end{equation}

Therefore, excluding normalization layers and the skip connection, the total number of trainable parameters in a SechKAN layer is
\begin{equation}
\boxed{
\begin{aligned}
\mathrm{SechKAN}_{\mathrm{param}}
&=
\mathrm{SechBasis}_{\mathrm{param}}
+\mathrm{Feature}_{\mathrm{param}}
+\mathrm{Grid}_{\mathrm{param}}
\\
&=
(G+2+\mathbb{I}_{\mathrm{width}})
+d_{\mathrm{in}}d_{\mathrm{out}}
+d_{\mathrm{out}}
+(G+1)
\\
&=
d_{\mathrm{in}}d_{\mathrm{out}}
+d_{\mathrm{out}}
+2G+3+\mathbb{I}_{\mathrm{width}}
\end{aligned}
}
\label{eq:sech_kan_no_norm_no_skip}
\end{equation}


For the implementation without the learnable width parameter $w$, the parameter count is $d_{\mathrm{in}}d_{\mathrm{out}} + d_{\mathrm{out}} + 2G + 3$. When the learnable width parameter is enabled, one additional parameter is introduced, resulting in $d_{\mathrm{in}}d_{\mathrm{out}} + d_{\mathrm{out}} + 2G + 4$. Thus, compared with an MLP layer, SechKAN introduces only a small number of additional parameters when excluding data normalization and the skip connection:
\begin{equation}
\mathrm{SechKAN}_{\mathrm{param}}
=
\mathrm{MLP}_{\mathrm{param}}
+2G+3+\mathbb{I}_{\mathrm{width}}
\label{eq:sech_kan_with_mlp}
\end{equation}
When $G\ll\min(d_{\mathrm{in}},d_{\mathrm{out}})$, this additional term is negligible compared with the feature projection term, resulting in a parameter complexity comparable to an MLP while retaining the nonlinear basis representation.

For data normalization, SechKAN has two normalization positions (Norm1 and Norm2). At each position, different normalization types can be applied, including LayerNorm, BatchNorm, Min--Max Norm, RMSNorm, or identity mapping. Each normalization type introduces a different number of learnable parameters. Min--Max normalization and identity mapping do not use any learnable parameters. For RMSNorm, each normalization layer contains only a learnable scale parameter, resulting in $d_{\mathrm{in}}$ trainable parameters per layer. Therefore, applying RMSNorm at both positions introduces a total of $2d_{\mathrm{in}}$ additional parameters. For LayerNorm and BatchNorm, each normalization layer contains a learnable weight parameter $\gamma$ and bias parameter $\beta$ for each feature dimension. Thus, both introduce $2d_{\mathrm{in}}$ trainable parameters per layer, resulting in $4d_{\mathrm{in}}$ additional parameters when applied at both normalization positions.

When the skip connection is enabled, it introduces an additional linear projection from $d_{\mathrm{in}}$ to $d_{\mathrm{out}}$, contributing $d_{\mathrm{in}}d_{\mathrm{out}} + d_{\mathrm{out}}$ trainable parameters. SechKAN obtains its maximum number of trainable parameters when LayerNorm or BatchNorm is applied at both normalization positions, the skip connection is enabled, and the learnable width parameter $w$ is included. The total number of trainable parameters is
\begin{equation}
\label{eq:sech+kan_full}
\begin{aligned}
\mathrm{SechKAN}_{\mathrm{param}}^{\mathrm{LN/BN+skip}}
&=
\mathrm{SechKAN}_{\mathrm{param}}^{\mathrm{max}} \\
&=
2d_{\mathrm{in}}d_{\mathrm{out}}
+2d_{\mathrm{out}}
+4d_{\mathrm{in}}
+2G + 4 \\
&= 
2\left(
(d_{\mathrm{out}}+2)(d_{\mathrm{in}}+1)
+G
\right)
\end{aligned}
\end{equation}

In the experiments, we use \Cref{eq:sech_kan_no_norm_no_skip} with the learnable width parameter \(w\) enabled and the skip connection disabled for function fitting. The reported parameter counts additionally include normalization parameters when applicable. For the selected SechKAN configurations used in the main Navier–Stokes, Shallow Water, and image-classification experiments, we use a SechKAN variant with LayerNorm applied at a single normalization position, while omitting the skip connection and the learnable width parameter $w$, resulting in a total number of trainable parameters of
\begin{equation}
\boxed{
\mathrm{SechKAN}_{\mathrm{param}}^{\mathrm{LN}}
=
d_{\mathrm{in}}d_{\mathrm{out}}
+d_{\mathrm{out}}
+2d_{\mathrm{in}}
+2G
+3
}
\label{eq:sech_kan_param_ln}
\end{equation}

\subsection{FLOPs for SechKAN, KAN, and MLP}

Let $d_{\text{in}}$ and $d_{\text{out}}$ denote the input and output dimensions of a layer, respectively. 
$G$ and $K$ represent the grid size and the spline order, which are specific hyperparameters in KAN-based architectures. 
Let $\alpha$ denote the FLOPs required for element-wise activation operations (e.g., ReLU, sigmoid, GELU, etc.) applied within each layer.

According to~\citet{yu2024kan}, the number of FLOPs in an MLP layer and a KAN layer are given:

\begin{equation}
\begin{aligned}
\text{MLP}_{\text{FLOP}} 
&= 2 d_{\text{in}}  d_{\text{out}}
+ \alpha d_{\text{out}}
\end{aligned}
\label{eq:flops_mlp}
\end{equation}

\begin{equation}
\begin{aligned}
\text{KAN}_{\text{FLOP}} 
&= \alpha d_{\text{in}} 
+ (d_{\text{in}} d_{\text{out}}) \times \Big[ 9K(G + 1.5K) + 2G - 2.5K + 3 \Big] \\
& 
\end{aligned}
\label{eq:flops_kan}
\end{equation}

The sech basis computation consists of subtraction, optional width scaling, hyperbolic cosine evaluation, reciprocal computation, element-wise scaling, and bias addition. Let $\mathbb{I}_{\mathrm{width}}$ denote an indicator variable, where $\mathbb{I}_{\mathrm{width}}=1$ when the learnable width parameter $w$ is enabled and $\mathbb{I}_{\mathrm{width}}=0$ otherwise. The subtraction operation contributes $d_{\mathrm{in}}G$ FLOPs, while the optional width scaling introduces an additional $\mathbb{I}_{\mathrm{width}}d_{\mathrm{in}}G$ FLOPs. The hyperbolic cosine evaluation contributes $d_{\mathrm{in}}G\beta$ FLOPs, where $\beta$ denotes the computational cost of the element-wise hyperbolic cosine function. The reciprocal operation $1/\cosh(\cdot)$ requires another $d_{\mathrm{in}}G$ FLOPs, and the element-wise scaling and bias addition together require $2d_{\mathrm{in}}G$ FLOPs. Therefore, the total computational cost of the sech basis computation is approximately $(4+\mathbb{I}_{\mathrm{width}} + \beta)d_{\mathrm{in}}G$ FLOPs.

Suppose that data normalization and skip connections are not used in SechKAN. The grid projection, which maps the $G$ basis responses to a single value, requires approximately $2d_{\mathrm{in}}G$ FLOPs. The feature projection from $d_{\mathrm{in}}$ to $d_{\mathrm{out}}$ requires approximately $2d_{\mathrm{in}}d_{\mathrm{out}}$ FLOPs, while the element-wise activation introduces an additional $d_{\mathrm{in}}\alpha$ FLOPs. Therefore, the total number of FLOPs of a SechKAN layer can be approximated as follows:

\begin{equation}
\boxed{
\begin{aligned}
\mathrm{SechKAN}_{\mathrm{FLOPs}}
&=
(4+\mathbb{I}_{\mathrm{width}} + \beta)d_{\mathrm{in}}G
+2d_{\mathrm{in}}G
+2d_{\mathrm{in}}d_{\mathrm{out}}
+d_{\mathrm{in}}\alpha \\
&=
d_{\mathrm{in}}
\left(
G(6+\mathbb{I}_{\mathrm{width}}+\beta)
+\alpha
+2d_{\mathrm{out}}
\right)
\end{aligned}
}
\label{eq:sechkan_flops}
\end{equation}

For data normalization, SechKAN supports two normalization positions (Norm1 and Norm2), where each position can employ LayerNorm, BatchNorm, Min--Max Norm, RMSNorm, or identity mapping. Each normalization strategy introduces a different computational overhead. Identity mapping introduces no additional computational overhead. Min--Max normalization computes the minimum and maximum values, followed by subtraction, division, scaling, and shifting, requiring approximately $5d_{\mathrm{in}}$ FLOPs per normalization layer. RMSNorm computes the root mean square, normalization, and learnable scaling, resulting in approximately $6d_{\mathrm{in}}$ FLOPs per layer. LayerNorm and BatchNorm additionally compute the mean and variance before normalization and affine transformation, requiring approximately $8d_{\mathrm{in}}$ FLOPs per layer during training. When normalization is applied at both Norm1 and Norm2, the total computational overhead becomes approximately $10d_{\mathrm{in}}$, $12d_{\mathrm{in}}$, and $16d_{\mathrm{in}}$ FLOPs for Min--Max Norm, RMSNorm, and LayerNorm/BatchNorm, respectively.

For skip connections, SechKAN optionally employs a residual branch that first applies an element-wise activation to the input features, followed by a linear projection from $d_{\mathrm{in}}$ to $d_{\mathrm{out}}$, and finally adds the projected features to the main branch. The activation operation introduces approximately $d_{\mathrm{in}}\alpha$ FLOPs, where $\alpha$ denotes the computational cost of the chosen activation function. The linear projection requires approximately $2d_{\mathrm{in}}d_{\mathrm{out}}$ FLOPs, and the residual addition contributes an additional $d_{\mathrm{out}}$ FLOPs. Therefore, enabling the skip connection introduces approximately $d_{\mathrm{in}}\alpha + 2d_{\mathrm{in}}d_{\mathrm{out}} + d_{\mathrm{out}}$ additional FLOPs. Since the activation is applied before the linear projection in the residual branch, its computational cost depends on the input dimension $d_{\mathrm{in}}$ rather than the output dimension $d_{\mathrm{out}}$.

SechKAN obtains the maximum computational complexity when both LayerNorm (or BatchNorm) and the skip connection are enabled. In this case, the total FLOPs are given by
\begin{equation}
\begin{aligned}
\mathrm{SechKAN}_{\mathrm{FLOP}}^{\mathrm{LN/BN+skip}}
&=
\mathrm{SechKAN}_{\mathrm{FLOP}}^{\max} \\
&= 
d_{in}\!\left(G(7+\beta) + \alpha + 2d_{out}\right)
+ 16d_{\mathrm{in}}
+d_{\mathrm{in}}\alpha
+2d_{\mathrm{in}}d_{\mathrm{out}}
+d_{\mathrm{out}} \\
&= d_{\mathrm{in}}\left(G(7+\beta)+2\alpha+4d_{\mathrm{out}}+16\right)+d_{\mathrm{out}}
\end{aligned}
\label{eq:sechkan_flops_max}
\end{equation}

In the experiments, we use \Cref{eq:sechkan_flops} with the learnable width parameter enabled ($\mathbb{I}_{\mathrm{width}}=1$) for function fitting. For the selected SechKAN configurations used in the main Navier–Stokes, Shallow Water, and image-classification experiments, we use a SechKAN variant with LayerNorm applied at a single normalization position, while omitting the skip connection and the learnable width parameter ($\mathbb{I}_{\mathrm{width}}=0$). Since LayerNorm introduces an additional computational overhead of $8d_{\mathrm{in}}$ FLOPs at a single normalization position, the total FLOPs are given by

\begin{equation}
\boxed{
\begin{aligned}
\mathrm{SechKAN}_{\mathrm{FLOPs}}^{\mathrm{LN}}
&=
d_{\mathrm{in}}
\left(
G(6+\beta)
+\alpha
+2d_{\mathrm{out}}
\right)
+8d_{\mathrm{in}} \\
&=
d_{\mathrm{in}}
\left(
G(6+\beta)
+\alpha
+2d_{\mathrm{out}}
+8
\right)
\end{aligned}
}
\label{eq:sechkan_flops_ln}
\end{equation}



\section{Experiments and Evaluation}
\label{sec:ex_eva}

We evaluate SechKAN against several KAN variants and MLPs on diverse tasks, including function fitting on 10 synthetic benchmark functions (1D--5D) and two representative AI Feynman equations, PDE surrogate modeling on the Navier--Stokes and Shallow Water datasets, and image classification on MNIST, Fashion-MNIST, CIFAR-10, and CIFAR-100. All experiments are conducted on a GeForce RTX 3060 Ti GPU. Fixed random seeds are used throughout all experiments to ensure reproducibility and consistent initialization. In particular, Python, NumPy, and PyTorch are seeded for all runs.


\subsection{Evaluation Metrics and Computational Complexity}

We evaluate model performance using standard task-specific metrics that reflect both accuracy and numerical fidelity.

For function fitting, we use the mean squared error (MSE):
\begin{equation}
\mathrm{MSE} = \frac{1}{N}\sum_{i=1}^{N}(y_i - \hat{y}_i)^2,
\end{equation}
where $N$ is the number of samples, $y_i$ is the ground-truth value, and $\hat{y}_i$ is the predicted value. MSE is widely used to measure pointwise regression accuracy and penalizes large deviations more strongly.

For PDE surrogate modeling, we use the relative $L2$ error:
\begin{equation}
\mathrm{Rel.\,L2} = \frac{\|u - \hat{u}\|_2}{\|u\|_2},
\end{equation}
where $u$ and $\hat{u}$ denote the exact and predicted solutions. This metric is scale-invariant and is commonly used to evaluate global solution accuracy in scientific computing.


For image classification, we report accuracy and macro-averaged F1-score. Accuracy is defined as the proportion of correctly classified samples:
\begin{equation}
\mathrm{Acc} = \frac{1}{N}\sum_{i=1}^{N}\mathbf{1}(\hat{y}_i = y_i),
\end{equation}
where $N$ is the number of samples, $y_i$ and $\hat{y}_i$ denote the true and predicted labels, respectively, and $\mathbf{1}(\cdot)$ is the indicator function. The macro-averaged F1-score is computed by first calculating the F1-score for each class:
\begin{equation}
\mathrm{F1}_c = \frac{2P_cR_c}{P_c + R_c},
\end{equation}
and then averaging the scores across all $C$ classes:
\begin{equation}
\mathrm{F1} = \frac{1}{C}\sum_{c=1}^{C}\mathrm{F1}_c,
\end{equation}
where $P_c$ and $R_c$ denote the precision and recall for class $c$, respectively.

To evaluate training efficiency, we define:
\begin{equation}
\mathrm{Acc/Time} = \frac{\mathrm{Acc}}{T}, \quad
\mathrm{Acc/Param} = \frac{\mathrm{Acc}}{P},
\end{equation}
where $T$ denotes training time and $P$ denotes the number of trainable parameters. These metrics capture the trade-off between predictive performance, computational cost, and model size.

Computational complexity is evaluated using the number of trainable parameters and forward-pass FLOPs (MFLOPs). Forward-pass FLOPs are measured using the \texttt{fvcore} profiler. Since several operators used in KAN-based models are not natively supported by \texttt{fvcore}, we implement custom operator handlers for unsupported element-wise operations, pooling layers, activation functions, and model-specific operators to ensure consistent FLOP counting across all compared methods. Linear and convolutional layers are counted using the default \texttt{fvcore} implementation. All models are profiled under identical input settings to ensure a fair comparison.

Computational efficiency is evaluated using peak GPU memory consumption (MB), forward and backward execution times, and training throughput. Peak GPU memory consumption is measured using \texttt{torch.cuda.max\_memory\_allocated()} during a single profiling epoch (the first epoch in our experiments). Forward and backward execution times are recorded using CUDA events after excluding the first 20 warm-up batches to avoid GPU initialization overhead and ensure steady-state execution. Training throughput is computed for each profiled batch as the number of processed samples divided by the sum of the forward and backward execution times, and the reported value is averaged across all remaining batches in the profiling epoch.

\subsection{Function Fitting}

\subsubsection{Visual Comparison of Model Predictions}
\begin{figure*}[!ht]
  \centering
\includegraphics[scale=0.45]{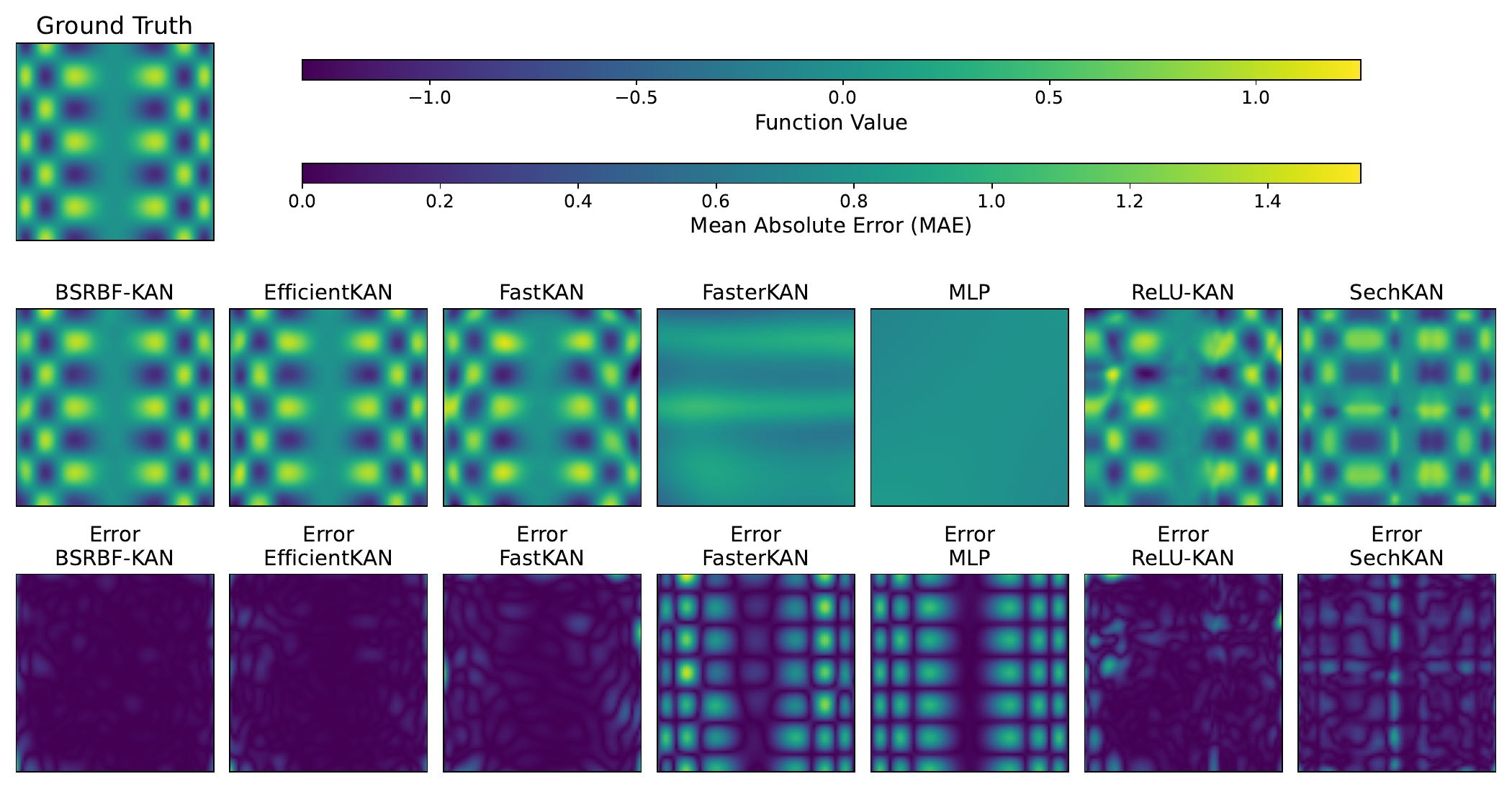}
  \centering
  \caption{Predictions and absolute error maps of different models compared with the ground truth during training on $f(x_1, x_2) = \sin(3 \pi x_1^2)\cos(3 \pi x_2)$. From visual inspection, BSRBF-KAN, EfficientKAN, and FastKAN closely match the ground truth, followed by ReLU-KAN and SechKAN.}
\label{fig:ff_output}
\end{figure*}

First, we conduct an experiment to evaluate the ability of each model to capture data features in a function fitting task. This experiment compares several KAN variants and an MLP baseline on a 2D nonlinear function,
$f(x_1, x_2) = \sin(3 \pi x_1^2)\cos(3 \pi x_2)$. This function was chosen due to its strong nonlinearity, partial periodicity, and the multiplicative interaction between the input variables. These characteristics make it a challenging yet representative benchmark for assessing a model's capacity to approximate highly nonlinear and coupled functional relationships. The selection of this function is not intended to favor any particular KAN variant, such as SechKAN.

The dataset, consisting of 600 samples, is divided into training and validation sets using an 8:2 ratio. The model achieving the lowest validation MSE is selected as the best-performing model. All models are trained for 2,000 epochs using the AdamW optimizer with a learning rate of $10^{-3}$, a weight decay of $10^{-4}$, and approximately 1,160 trainable parameters. This network configuration is used only for the visualization experiment in \Cref{fig:ff_output}.



Performance is evaluated by comparing the ground truth, model predictions, and corresponding error maps. For SechKAN, we employ Min--Max normalization in Norm1, use the SELU activation function, and set the width parameter to \texttt{w}. As shown in \Cref{fig:ff_output}, FastKAN, EfficientKAN, BSRBF-KAN, ReLU-KAN, and SechKAN are selected for comparison in the function-fitting task, while the remaining models are omitted due to their poorer performance.


\subsubsection{Quantitative Evaluation on Function Fitting}
\label{sec:ex_function_fitting}

\begin{table*}[!tb]
\centering
\caption{Functions and network structures used in the experiments. The network structures were designed so that all models have approximately the same number of trainable parameters. ``Other KANs'' refers to BSRBF-KAN, EfficientKAN, and FastKAN. ReLU-KAN uses the same network structures except for $f_{10}$, where the architecture is $[5,9,9,1]$.}
\begin{tabular}{p{9.5cm}cc}
\hline
\textbf{Functions} & \textbf{Other KANs} & \textbf{SechKAN} \\
\hline

$f_1(x)=\exp(-4x^2)\sin(6\pi x)$
& [1,1] & [1,1] \\ \hline

$f_2(x)=\sin(4\pi x^2)\cos(2\pi x)$
& [1,1] & [1,1] \\ \hline

$f_3(x_1,x_2)=\sin(3\pi x_1^2)\cos(3\pi x_2)$
& [2,4,4,1] & [2,14,14,1] \\ \hline

$f_4(x_1,x_2,x_3)=
\exp\!\left(-3(x_1^2+x_2^2)\right)
\sin\!\left(4\pi x_3+x_1x_2\right)$
& [3,4,4,1] & [3,16,16,1] \\ \hline

$\begin{aligned}
f_5(x_1,x_2,x_3,x_4)
&=\sin\!\big(3\pi(x_1x_2+x_3)\big)\\
&\quad\times\cos\!\big(2\pi(x_2x_4)\big)
\end{aligned}$
& [4,8,8,1] & [4,32,32,1] \\ \hline

$f_6(x_1,x_2,x_3)
=
x_1\left(1+x_2\cos x_3\right)$
& [3,4,4,1] & [3,16,16,1] \\ \hline

$f_7(x_1,x_2,x_3,x_4)
=
\frac12x_1\left(x_2^2+x_3^2+x_4^2\right)$
& [4,8,8,1] & [4,32,32,1] \\ \hline

$f_8(x_1,x_2)=
\sin(20\pi x_1)\cos(18\pi x_2)$
& [2,4,4,1] & [2,14,14,1] \\ \hline

$f_9(x_1,x_2,x_3)=
\tanh\!\left(50(x_1-x_2+0.5x_3-0.5)\right)$
& [3,4,4,1] & [3,16,16,1] \\ \hline

$\begin{aligned}
f_{10}(x_1,x_2,x_3,x_4,x_5)
&=
\frac{
\sin(3\pi I)
+0.5\cos\!\left(2\pi(x_1+x_3-x_5)\right)
}{
1+\sum_{i=1}^{5}x_i^2
},\\
I&=x_1x_2+x_2x_3+x_3x_4+x_4x_5+x_5x_1
\end{aligned}$
& [5,8,8,1] & [5,32,32,1] \\ \hline

\hline
\label{tab:function_list}
\end{tabular}
\end{table*}

\begin{table*}[!tb]
\centering
\caption{Average performance comparison of models on 10 functions ($f_1$–$f_10$) and 10 seeds. For each model, the number of parameters used, the MSE loss, and the running time are computed as the average over 100 runs (10 functions × 10 seeds).}
\begin{tabular}{p{2.8cm}p{2cm}p{2.2cm}p{2cm}p{2cm}}
\hline
\textbf{Model} & \textbf{Function} & \textbf{Used Params} & \textbf{Loss (MSE) $\downarrow$} & \textbf{Time (s) $\downarrow$} \\ \hline

BSRBF-KAN     & Avg & 648 & $3.22\times10^{-2}$ & 6.89 \\
EfficientKAN  & Avg & 648 & $1.93\times10^{-2}$ & 6.69 \\
FastKAN       & Avg & 658 & $4.33\times10^{-2}$ & \textbf{5.65} \\
ReLU-KAN      & Avg & 643 & $3.85\times10^{-2}$ & 5.73 \\
SechKAN       & Avg & 651 & $\mathbf{1.75\times10^{-2}}$ & 6.74 \\
\hline

\end{tabular}
\label{tab:ff_average}
\end{table*}

\begin{figure*}[htbp]
  \centering
\includegraphics[scale=0.52]{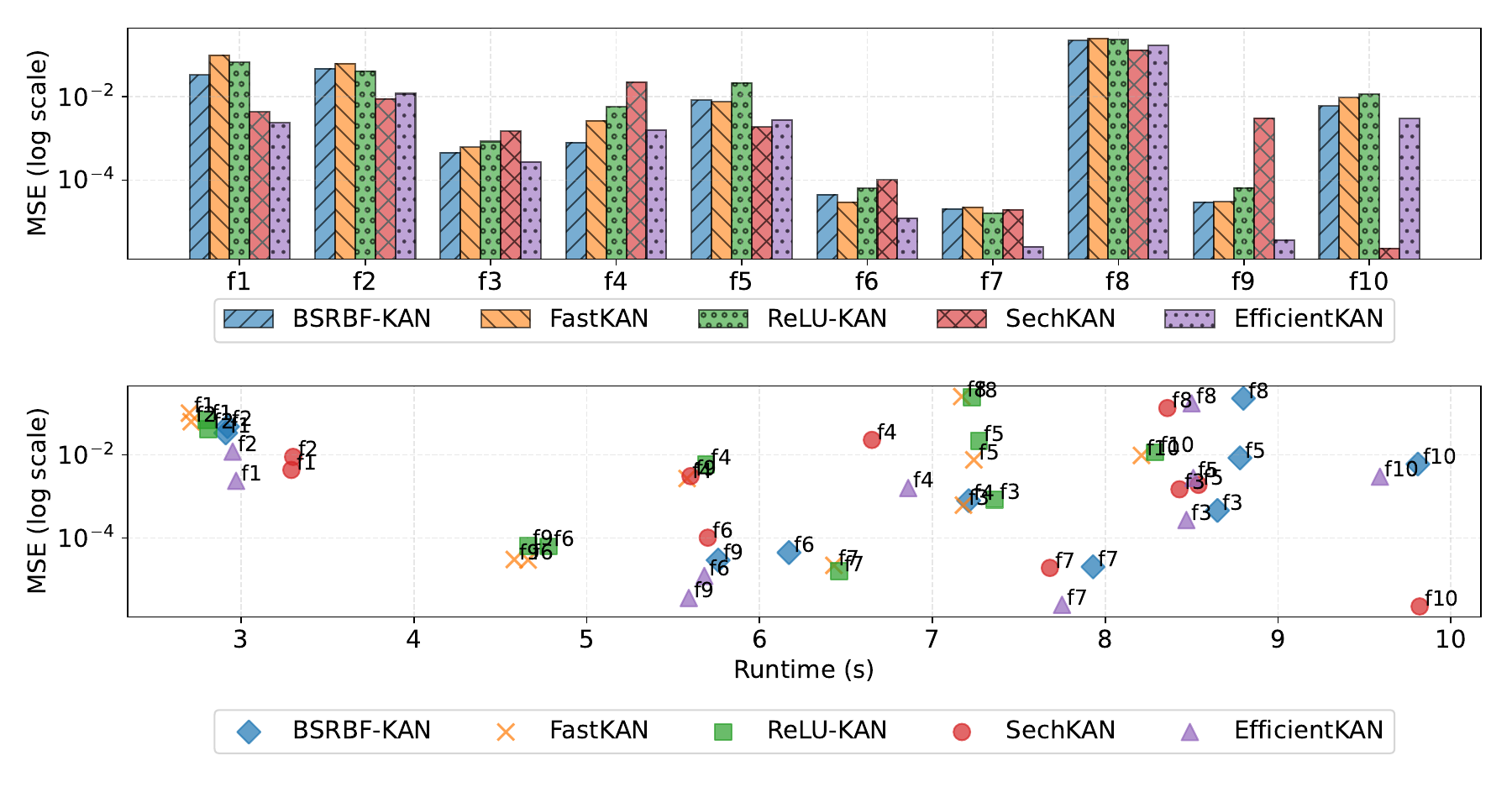}
  \centering
  \caption{Performance comparison of MSE loss and runtime for different models on 10 functions ($f_1$–$f_10$) over 10 seeds (0–9). The reported MSE values are averaged across seeds for each function and model. (Top) MSE of the models for each function (log scale). (Bottom) Relationship between log-scale MSE and runtime across all models and functions.}
\label{fig:ff_plot}
\end{figure*}

\begin{table*}[!ht]
\centering
\caption{Per-function permutation test results comparing SechKAN with baseline models ($n=10$ functions). The permutation test is performed on the mean MSE of each function. Mean difference is computed as SechKAN minus the baseline.}
\begin{tabular}{p{4cm}p{3cm}p{3cm}p{2.5cm}}
\hline
\textbf{Comparison} & \textbf{Mean Difference} $\downarrow$ & \textbf{$p$-value} $\downarrow$ & \textbf{n Functions} \\
\hline

SechKAN vs FastKAN      & -0.025807 & 0.152600 & 10 \\
SechKAN vs ReLU-KAN     & -0.021036 & 0.110100 & 10 \\
SechKAN vs BSRBF-KAN    & -0.014759 & 0.152600 & 10 \\
SechKAN vs EfficientKAN & -0.001836 & 0.747600 & 10 \\

\hline
\end{tabular}
\label{tab:func_level_test}
\end{table*}

\begin{table*}[!ht]
\centering
\caption{Seed-level permutation test results for each function ($n=10$ seeds per function). The permutation test is performed on paired MSE differences over the 10 random seeds. Mean difference is computed as SechKAN minus the baseline.}
\begin{tabular}{p{2cm}p{4cm}p{3cm}p{2.5cm}p{2cm}}
\hline
\textbf{Function} & \textbf{Comparison} & \textbf{Mean Difference $\downarrow$} & \textbf{$p$-value} & \textbf{n Seeds} \\
\hline

\multirow{4}{2cm}{\textbf{$f_1$}}
& SechKAN vs EfficientKAN &  0.001945 & 0.1375 & 10 \\
& SechKAN vs FastKAN      & -0.095961 & 0.0015 & 10 \\
& SechKAN vs BSRBF-KAN    & -0.028931 & 0.0015 & 10 \\
& SechKAN vs ReLU-KAN     & -0.063311 & 0.0015 & 10 \\
\hline

\multirow{4}{2cm}{\textbf{$f_2$}}
& SechKAN vs EfficientKAN & -0.003059 & 0.2158 & 10 \\
& SechKAN vs FastKAN      & -0.053119 & 0.0015 & 10 \\
& SechKAN vs BSRBF-KAN    & -0.038549 & 0.0015 & 10 \\
& SechKAN vs ReLU-KAN     & -0.032209 & 0.0015 & 10 \\
\hline

\multirow{4}{2cm}{\textbf{$f_3$}}
& SechKAN vs EfficientKAN &  0.001209 & 0.0713 & 10 \\
& SechKAN vs FastKAN      &  0.000862 & 0.2468 & 10 \\
& SechKAN vs BSRBF-KAN    &  0.001020 & 0.1352 & 10 \\
& SechKAN vs ReLU-KAN     &  0.000633 & 0.3838 & 10 \\
\hline

\multirow{4}{2cm}{\textbf{$f_4$}}
& SechKAN vs EfficientKAN &  0.021145 & 0.0578 & 10 \\
& SechKAN vs FastKAN      &  0.020044 & 0.0600 & 10 \\
& SechKAN vs BSRBF-KAN    &  0.021944 & 0.0500 & 10 \\
& SechKAN vs ReLU-KAN     &  0.016958 & 0.1547 & 10 \\
\hline

\multirow{4}{2cm}{\textbf{$f_5$}}
& SechKAN vs EfficientKAN & -0.000889 & 0.3121 & 10 \\
& SechKAN vs FastKAN      & -0.005662 & 0.0015 & 10 \\
& SechKAN vs BSRBF-KAN    & -0.006491 & 0.0015 & 10 \\
& SechKAN vs ReLU-KAN     & -0.019432 & 0.0015 & 10 \\
\hline

\multirow{4}{2cm}{\textbf{$f_6$}}
& SechKAN vs EfficientKAN &  0.000090 & 0.0015 & 10 \\
& SechKAN vs FastKAN      &  0.000072 & 0.0052 & 10 \\
& SechKAN vs BSRBF-KAN    &  0.000057 & 0.0473 & 10 \\
& SechKAN vs ReLU-KAN     &  0.000039 & 0.2588 & 10 \\
\hline

\multirow{4}{2cm}{\textbf{$f_7$}}
& SechKAN vs EfficientKAN &  0.000017 & 0.0015 & 10 \\
& SechKAN vs FastKAN      & -0.000003 & 0.6426 & 10 \\
& SechKAN vs BSRBF-KAN    & -0.000001 & 0.8954 & 10 \\
& SechKAN vs ReLU-KAN     &  0.000003 & 0.3468 & 10 \\
\hline

\multirow{4}{2cm}{\textbf{$f_8$}}
& SechKAN vs EfficientKAN & -0.038902 & 0.2417 & 10 \\
& SechKAN vs FastKAN      & -0.117702 & 0.0034 & 10 \\
& SechKAN vs BSRBF-KAN    & -0.093702 & 0.0336 & 10 \\
& SechKAN vs ReLU-KAN     & -0.104602 & 0.0123 & 10 \\
\hline

\multirow{4}{2cm}{\textbf{$f_9$}}
& SechKAN vs EfficientKAN &  0.003055 & 0.0015 & 10 \\
& SechKAN vs FastKAN      &  0.003028 & 0.0015 & 10 \\
& SechKAN vs BSRBF-KAN    &  0.003029 & 0.0015 & 10 \\
& SechKAN vs ReLU-KAN     &  0.002994 & 0.0080 & 10 \\
\hline

\multirow{4}{2cm}{\textbf{$f_{10}$}}
& SechKAN vs EfficientKAN & -0.002968 & 0.0015 & 10 \\
& SechKAN vs FastKAN      & -0.009634 & 0.0015 & 10 \\
& SechKAN vs BSRBF-KAN    & -0.005962 & 0.0015 & 10 \\
& SechKAN vs ReLU-KAN     & -0.011432 & 0.0015 & 10 \\
\hline

\end{tabular}
\label{tab:seed_level_test}
\end{table*}

Next, we evaluate 10 functions with input dimensions ranging from 1D to 5D. \Cref{tab:function_list} lists the functions and their network structures. Functions $f_1$ and $f_2$ are one-dimensional functions with localized and warped oscillations. Functions $f_3$--$f_5$ increase the input dimension and include anisotropic patterns, localized nonlinear interactions, and coupled nonlinear terms. Functions $f_6$ and $f_7$ are adopted from the AI Feynman dataset~\cite{udrescu2020ai}, corresponding to Equation III.17.37 (angular distribution in weak decay) and Equation I.13.4 (kinetic energy), respectively. Function $f_8$ is a high-frequency oscillatory function, $f_9$ is a near-discontinuous function with a sharp transition, and $f_{10}$ is a five-dimensional function with multiple nonlinear interactions, making it the most challenging function in our experiments.

Based on \Cref{fig:ff_output}, we choose SechKAN, BSRBF-KAN, FastKAN, EfficientKAN, and ReLU-KAN for the experiment. All models are trained for 500 epochs using the architectures specified in \Cref{tab:function_list}, with comparable parameter counts across models to ensure a fair comparison. For training, we use the Adam optimizer with a learning rate of $10^{-2}$ and the mean squared error (MSE) loss. No learning rate scheduler is used.

ReLU-KAN employs ReLU, EfficientKAN and FastKAN use SiLU, whereas SechKAN uses SELU. Moreover, SechKAN applies Min-Max and BatchNorm in Norm1 in each layer, introduces the learnable width parameter $w$ (\Cref{eq:sech_basis}), and removes the skip connection (\texttt{use\_base\_update=False}). SechKAN adopts grid resolutions of 4 or 8 depending on the complexity of each benchmark function, while the baseline models adjust the grid size, number of grids, or spline order for each function to maintain a comparable parameter budget.

As shown in \Cref{tab:ff_average} and \Cref{fig:ff_plot}, SechKAN achieves the lowest average MSE while maintaining a parameter count comparable to the other models. It consistently delivers strong approximation performance across the benchmark suite, achieving the best results on $f_2$, $f_5$, $f_8$, and $f_{10}$, while remaining competitive on $f_1$, $f_3$, $f_6$, and $f_7$. Although EfficientKAN and BSRBF-KAN outperform SechKAN on a few individual functions, SechKAN achieves the best overall approximation accuracy when averaged over all functions and random seeds. These results suggest that SechKAN is particularly effective for highly nonlinear functions involving oscillatory patterns, multiplicative interactions, and higher-dimensional nonlinear coupling. In terms of runtime, SechKAN requires only slightly more training time than EfficientKAN, whereas FastKAN is the fastest but exhibits the highest average MSE. Additional experimental details, including the benchmark functions, model configurations, random seeds, and complete results, are provided in \Cref{sec:ff_appendix}.

To further evaluate the performance of SechKAN, we employ permutation tests as a non-parametric method for assessing statistical significance under both function-level and seed-level settings. This analysis examines whether the observed MSE differences between SechKAN and the baseline models are statistically significant. As shown in \Cref{tab:func_level_test}, the function-level analysis indicates that SechKAN achieves lower average MSE than all baseline models, yielding consistently negative mean differences. However, none of the corresponding $p$-values are below 0.05, suggesting that these improvements are not statistically significant, likely due to the limited sample size of only 10 benchmark functions.

In contrast, the seed-level analysis (\Cref{tab:seed_level_test}), which evaluates each function over 10 random seeds, provides a more detailed comparison. SechKAN significantly outperforms FastKAN, BSRBF-KAN, and ReLU-KAN on several functions, particularly $f_1$, $f_2$, $f_5$, $f_8$, and $f_{10}$. The comparison between SechKAN and EfficientKAN is more balanced: no statistically significant differences are observed on six of the ten functions. EfficientKAN performs significantly better on $f_6$, $f_7$, and $f_9$, whereas SechKAN performs significantly better only on $f_{10}$.

In summary, the permutation tests indicate that although the function-level comparison does not establish statistical significance, the seed-level analysis provides stronger evidence that SechKAN achieves competitive performance across a diverse set of benchmark functions, with statistically significant improvements on several functions under different random initializations.

\subsection{PDE surrogate modeling}

\subsubsection{Dataset Generation}


In this section, we perform PDE surrogate modeling. To construct the datasets, the Navier--Stokes and Shallow Water equations are solved using two-dimensional pseudo-spectral solvers with a fourth-order Runge--Kutta (RK4) time integration scheme. For the Navier--Stokes problem, the simulation models viscous fluid dynamics from a random initial vorticity field with external forcing, producing coordinate-solution pairs that map the spatiotemporal coordinates $(x, y, t)$ to the velocity components $(u, v)$. For the Shallow Water problem, the simulation evolves the water height and velocity fields under gravitational and viscous effects on a structured spatial grid with a fixed Gaussian--sinusoidal bathymetry, producing coordinate-solution pairs that map $(x, y, t)$ to $(h, u, v)$. Throughout the simulations, the physical parameters (e.g., viscosity, gravity, bathymetry, and forcing) remain fixed, resulting in a single simulation trajectory for each PDE. The detailed simulation configurations are reported in \Cref{sec:app_ex_detail_pde}.

\begin{figure*}[htbp]
  \centering
\includegraphics[scale=0.40]{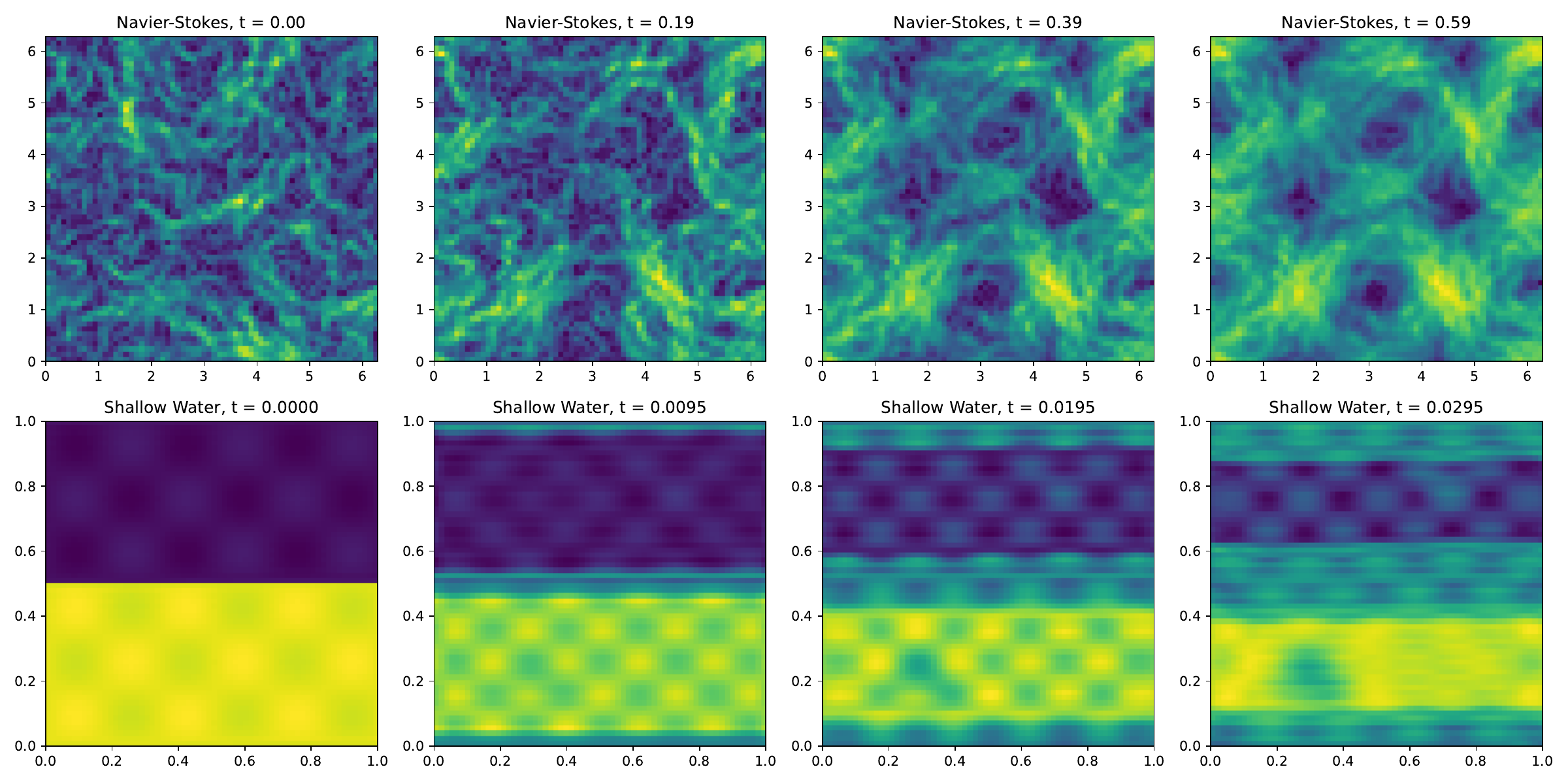}
  \centering
  \caption{Visualization of the generated Navier–Stokes and Shallow Water datasets at different time steps. The first row shows the evolution of velocity magnitude in the Navier–-Stokes simulation, where small-scale turbulent structures gradually merge into larger coherent flow patterns. The second row shows the evolution of water surface height in the Shallow Water simulation, exhibiting wave propagation and layer dynamics over time.}
  
\label{fig:pde_plot}
\end{figure*}

Thus, the task is to learn the continuous mappings $(x, y, t)\rightarrow(u, v)$ for the Navier--Stokes problem and $(x, y, t)\rightarrow(h, u, v)$ for the Shallow Water problem within a single simulation trajectory, rather than learning an operator that generalizes across multiple PDE instances with different initial conditions, boundary conditions, forcing terms, or physical parameters. The resulting coordinate-solution pairs are divided into training, validation, and testing sets using a time-indexing split with a ratio of 7:1.5:1.5, followed by normalization based on the training data statistics. \Cref{fig:pde_plot} presents samples from the generated PDE datasets at different time steps.

\subsubsection{Experimental Setup}

For both the Navier--Stokes and Shallow Water experiments, all models are trained using the AdamW optimizer with a learning rate of $10^{-3}$ and a weight decay factor of $10^{-4}$. The mean squared error (MSE) loss function is used during training, while a ReduceLROnPlateau scheduler and early stopping with a patience of 20 epochs are employed to improve training stability and prevent overfitting. 

The models take the spatial-temporal coordinates $(x, y, t)$ as inputs and predict the corresponding physical quantities of each PDE problem. The performance of each model is evaluated using RMSE and relative L2 error on the training, validation, and testing sets. All experiments are trained for a maximum of 100 epochs and repeated over five different random seeds. The final results are reported as the mean and standard deviation. Furthermore, the training time, testing time, and number of trainable parameters are recorded for efficiency analysis. 

All models are implemented with two hidden layers of comparable network capacity to ensure a fair evaluation with approximately the same number of trainable parameters. The MLP and SechKAN models use wider hidden dimensions of 196 and 192, respectively, whereas the other KAN variants employ approximately 60--70 neurons to achieve comparable parameter counts. For spline-based KAN models, including EfficientKAN and BSRBF-KAN, the spline order is set to $3$ with a grid size of $5$. FastKAN and FasterKAN use $8$ grid points, while ReLU-KAN employs a grid size of $5$ with kernel parameter $k=3$.

For SechKAN, data normalization is applied to all hidden layers except the input layer, and the SiLU activation function is used as the base activation. In addition, the skip connection (\texttt{use\_base\_update=False}) and the width parameter option (\texttt{use\_width=False}) are disabled to focus on the effect of the hyperbolic secant basis functions. For the PDE surrogate modeling experiments, LayerNorm is applied at Norm1 for both datasets, with the number of grids set to 4 for Navier--Stokes and 8 for Shallow Water. These configurations are selected based on the best model performance observed in experiments, with further details, including extra experiments, provided in \Cref{sec:app_ex_detail_pde}.

\subsubsection{Experimental Results}

\begin{table}[ht]
\centering
\caption{Performance comparison of SechKAN variants and baseline models on the Navier--Stokes dataset. SechKAN uses \texttt{num\_grids = 4}. Results are reported as mean $\pm$ standard deviation across 5 random seeds.}
\begin{tabular}{lcccc}
\hline
\textbf{Model} &
\textbf{Val. Rel. L2 $\downarrow$} &
\textbf{Test Rel. L2 $\downarrow$} &
\textbf{Runtime (s) $\downarrow$} &
\textbf{Used Params} \\
\hline
BSRBF-KAN &
0.09664 $\pm$ 0.00728 &
0.30327 $\pm$ 0.01853 &
422.67 $\pm$ 3.29 &
39744 \\
EfficientKAN &
\textbf{0.08496 $\pm$ 0.00345} &
\textbf{0.26524 $\pm$ 0.01692} &
388.91 $\pm$ 2.15 &
39630 \\
FastKAN &
0.30503 $\pm$ 0.01253 &
0.72194 $\pm$ 0.02485 &
\textbf{86.05 $\pm$ 1.42} &
38684 \\
FasterKAN &
0.26852 $\pm$ 0.06415 &
0.55705 $\pm$ 0.13917 &
179.68 $\pm$ 51.91 &
39152 \\


MLP &
0.69984 $\pm$ 0.00096 &
0.72133 $\pm$ 0.00361 &
130.69 $\pm$ 21.11 &
39396 \\

ReLU-KAN &
0.78003 $\pm$ 0.01123 &
0.89849 $\pm$ 0.03682 &
139.67 $\pm$ 39.35 &
39782 \\
SechKAN &
0.13664 $\pm$ 0.01529 &
0.27695 $\pm$ 0.00954 &
266.04 $\pm$ 1.87 &
39011 \\

\hline
\end{tabular}
\label{tab:navier_stoke_comparison}
\end{table}

\begin{table}[ht]
\centering
\caption{Performance comparison of SechKAN and baseline models on the Shallow Water dataset. SechKAN uses \texttt{num\_grids = 8} and LayerNorm at Norm1. Results are reported as mean $\pm$ standard deviation over 5 random seeds.}
\begin{tabular}{lcccc}
\hline
\textbf{Model} &
\textbf{Val. Rel. L2 $\downarrow$} &
\textbf{Test Rel. L2 $\downarrow$} &
\textbf{Runtime (s) $\downarrow$} &
\textbf{Used Params} \\
\hline

BSRBF-KAN &
0.15825 $\pm$ 0.01704 &
0.41038 $\pm$ 0.05549 &
215.60 $\pm$ 119.96 &
40320 \\

EfficientKAN &
0.20637 $\pm$ 0.01049 &
0.57372 $\pm$ 0.03928 &
143.23 $\pm$ 40.57 &
40230 \\

FastKAN &
0.21505 $\pm$ 0.01030 &
0.59755 $\pm$ 0.04715 &
116.11 $\pm$ 40.31 &
39252 \\

FasterKAN &
0.22423 $\pm$ 0.02903 &
0.39949 $\pm$ 0.03972 &
178.42 $\pm$ 65.66 &
39688 \\

MLP &
0.41146 $\pm$ 0.00383 &
0.46458 $\pm$ 0.00813 &
\textbf{108.28 $\pm$ 69.98} &
38016 \\

ReLU-KAN &
0.41802 $\pm$ 0.01192 &
0.61521 $\pm$ 0.00488 &
700.64 $\pm$ 92.31 &
40311 \\

\textbf{SechKAN} &
\textbf{0.12637 $\pm$ 0.00567} &
\textbf{0.29879 $\pm$ 0.01425} &
196.59 $\pm$ 18.88 &
39228 \\


\hline
\end{tabular}
\label{tab:shallow_water_comparison}
\end{table}

\Cref{tab:navier_stoke_comparison} and \Cref{tab:shallow_water_comparison} show the validation and test Rel L2 scores, training times, and parameter counts of SechKAN, existing KAN variants, and MLPs. On the Navier--Stokes dataset, EfficientKAN achieves the lowest validation and test Rel L2 scores. However, SechKAN attains comparable predictive performance while reducing training time by approximately 31.6\%. On the Shallow Water dataset, SechKAN achieves the best validation and test Rel L2 scores among all evaluated models. Compared with the second-best model, FasterKAN, SechKAN requires only 10.2\% more training time while achieving substantially better prediction accuracy.

SechKAN also demonstrates consistent performance across random seeds. On the Navier--Stokes dataset, its test Rel L2 standard deviation is lower than those of EfficientKAN, BSRBF-KAN, FastKAN, and FasterKAN, indicating lower sensitivity to random initialization. A similar pattern is observed on the Shallow Water dataset, where SechKAN achieves the lowest test Rel L2 standard deviation among all KAN variants. Moreover, SechKAN exhibits stable training times. On the Navier--Stokes dataset, its training-time standard deviation is only 1.87 seconds, comparable to the most stable baselines and substantially lower than those of FasterKAN, MLP, and ReLU-KAN. On the Shallow Water dataset, SechKAN records the lowest training-time standard deviation among all evaluated models, at 18.88 seconds. In short, these results indicate that SechKAN provides a stable training process while maintaining competitive predictive performance and computational efficiency.

\subsection{Image classification}

In this experiment, we evaluate SechKAN, MLP, CNN, and SechKAN-CNN on four image classification datasets: MNIST, Fashion-MNIST, CIFAR-10, and CIFAR-100. All models are trained for 10 epochs using the AdamW optimizer with a OneCycleLR scheduler and SiLU activation. The OneCycleLR scheduler is adopted to accelerate convergence and reduce training time while maintaining or improving performance. The training procedure is controlled by key hyperparameters, including the learning rate (default $10^{-3}$) and weight decay (default $10^{-4}$).
Each configuration is repeated with 5 random seeds (0, 1, 2, 3, 4), and results are reported as average values to ensure robustness. For a fair comparison, all network architectures are designed to have comparable numbers of parameters, as summarized in \Cref{tab:ic_model_params}.

\begin{table*}[ht]
\centering
\caption{Model configurations, parameter counts, MFLOPs, and peak GPU memory used in the image classification experiments.}
\begin{tabular}{p{2.2cm}p{2.2cm}p{4.9cm}p{1.2cm}p{1.2cm}p{1.8cm}}
\hline
\textbf{Dataset} & \textbf{Model} & \textbf{Network structure} & \textbf{Used Params} & \textbf{MFLOPs $\downarrow$} & \textbf{Peak Mem. (MB) $\downarrow$}\\
\hline

\multirow{4}{*}{\textbf{\shortstack[l]{MNIST\\Fashion-MNIST}}}
& SechKAN      & (784, 64, 10)                  & 52,608  & 0.096 & 23.84 \\
& MLP          & (784, 64, 10)                  & 50,816  & \textbf{0.051} & \textbf{17.42} \\
& CNN          & 2 Conv layers + 2 MLP layers   & 52,138  & 0.378 & 19.87 \\
& SechKAN-CNN & 2 Conv layers + 2 SechKAN layers & 52,160 & 0.418 & 21.09 \\
\hline

\multirow{4}{*}{\textbf{CIFAR-10}}
& SechKAN      & (3072, 64, 10)                 & 203,632 & 0.501 & 31.84 \\
& MLP          & (3072, 64, 10)                 & 197,248 & \textbf{0.198} & \textbf{20.21} \\
& CNN          & 2 Conv layers + 2 MLP layers   & 196,078 & 3.678 & 28.10 \\
& SechKAN-CNN & 2 Conv layers + 2 SechKAN layers & 196,100 & 3.828 & 32.76 \\
\hline

\multirow{4}{*}{\textbf{CIFAR-100}}
& SechKAN      & (3072, 256, 100)               & 819,082 & 1.135 & 43.58 \\
& MLP          & (3072, 256, 100)               & 812,032 & \textbf{0.813} & \textbf{32.70} \\
& CNN          & 2 Conv layers + 2 MLP layers   & 804,438 & 4.287 & 36.79 \\
& SechKAN-CNN & 2 Conv layers + 2 SechKAN layers & 804,460 & 4.445 & 43.70 \\
\hline

\end{tabular}
\label{tab:ic_model_params}
\end{table*}

\begin{table*}[!htbp]
\centering
\caption{Performance comparison of MLP, SechKAN, CNN, and SechKAN-CNN across datasets. Results are averaged over five runs, with one run per seed (0, 1, 2, 3, and 4).}
\begin{tabular}{p{2.4cm}p{2.4cm}p{2cm}p{2cm}p{2cm}p{2.8cm}}
\hline
\textbf{Dataset} & \textbf{Network} & \textbf{Train. Acc. $\uparrow$} & \textbf{Val. Acc. $\uparrow$} & \textbf{Val. F1 $\uparrow$} & \textbf{Runtime (s) $\downarrow$} \\
\hline

\multirow{4}{*}{MNIST}
& MLP & $97.97 \pm 0.02$ & $97.28 \pm 0.05$ & $97.24 \pm 0.05$ & $69.59 \pm 5.22$ \\
& SechKAN & $99.52 \pm 0.07$ & $97.68 \pm 0.08$ & $97.65 \pm 0.08$ & $70.73 \pm 2.32$ \\
& SechKAN-CNN & $99.97 \pm 0.01$ & $98.94 \pm 0.04$ & $98.94 \pm 0.04$ & $179.76 \pm 13.41$ \\
& CNN & $99.95 \pm 0.00$ & $\mathbf{99.11 \pm 0.02}$ & $\mathbf{99.10 \pm 0.02}$ & $132.52 \pm 5.42$ \\
\hline

\multirow{4}{*}{Fashion-MNIST}
& MLP & $90.57 \pm 0.01$ & $87.93 \pm 0.04$ & $87.85 \pm 0.04$ & $67.69 \pm 5.63$ \\
& SechKAN & $92.82 \pm 0.06$ & $89.03 \pm 0.07$ & $88.95 \pm 0.07$ & $70.69 \pm 4.25$ \\
& SechKAN-CNN & $95.41 \pm 0.21$ & $90.47 \pm 0.06$ & $90.44 \pm 0.07$ & $179.18 \pm 13.30$ \\
& CNN & $94.86 \pm 0.08$ & $\mathbf{91.05 \pm 0.06}$ & $\mathbf{91.03 \pm 0.06}$ & $135.60 \pm 4.53$ \\
\hline

\multirow{4}{*}{CIFAR-10}
& MLP & $61.18 \pm 0.15$ & $51.69 \pm 0.24$ & $51.35 \pm 0.24$ & $111.98 \pm 5.81$ \\
& SechKAN & $64.46 \pm 0.91$ & $52.59 \pm 0.15$ & $52.38 \pm 0.15$ & $132.23 \pm 8.70$ \\
& SechKAN-CNN & $88.54 \pm 0.82$ & $\mathbf{71.66 \pm 0.20}$ & $\mathbf{71.63 \pm 0.19}$ & $178.78 \pm 7.45$ \\
& CNN & $93.92 \pm 0.61$ & $70.72 \pm 0.19$ & $70.71 \pm 0.19$ & $116.49 \pm 3.50$ \\
\hline

\multirow{4}{*}{CIFAR-100}
& MLP & $44.03 \pm 0.13$ & $23.94 \pm 0.12$ & $23.17 \pm 0.14$ & $108.71 \pm 9.00$ \\
& SechKAN & $55.56 \pm 2.24$ & $24.20 \pm 2.16$ & $23.65 \pm 2.10$ & $140.89 \pm 2.50$ \\
& SechKAN-CNN & $79.97 \pm 3.54$ & $\mathbf{41.94 \pm 0.31}$ & $\mathbf{41.74 \pm 0.30}$ & $181.80 \pm 1.99$ \\
& CNN & $98.18 \pm 0.20$ & $39.61 \pm 0.25$ & $39.77 \pm 0.24$ & $116.95 \pm 2.44$ \\

\hline
\multirow{4}{*}{Average}
& MLP & 73.44 & 65.21 & 64.90 & 89.49 \\
& SechKAN & 78.09 & 65.88 & 65.66 & 103.64 \\
& SechKAN-CNN & 90.97 & $\mathbf{75.75}$ & $\mathbf{75.69}$ & 179.88 \\
& CNN & $\mathbf{96.73}$ & 75.12 & 75.15 & $\mathbf{125.39}$ \\
\hline

\end{tabular}
\label{tab:ic_result}
\end{table*}

\begin{figure*}[htbp]
  \centering
\includegraphics[scale=0.52]{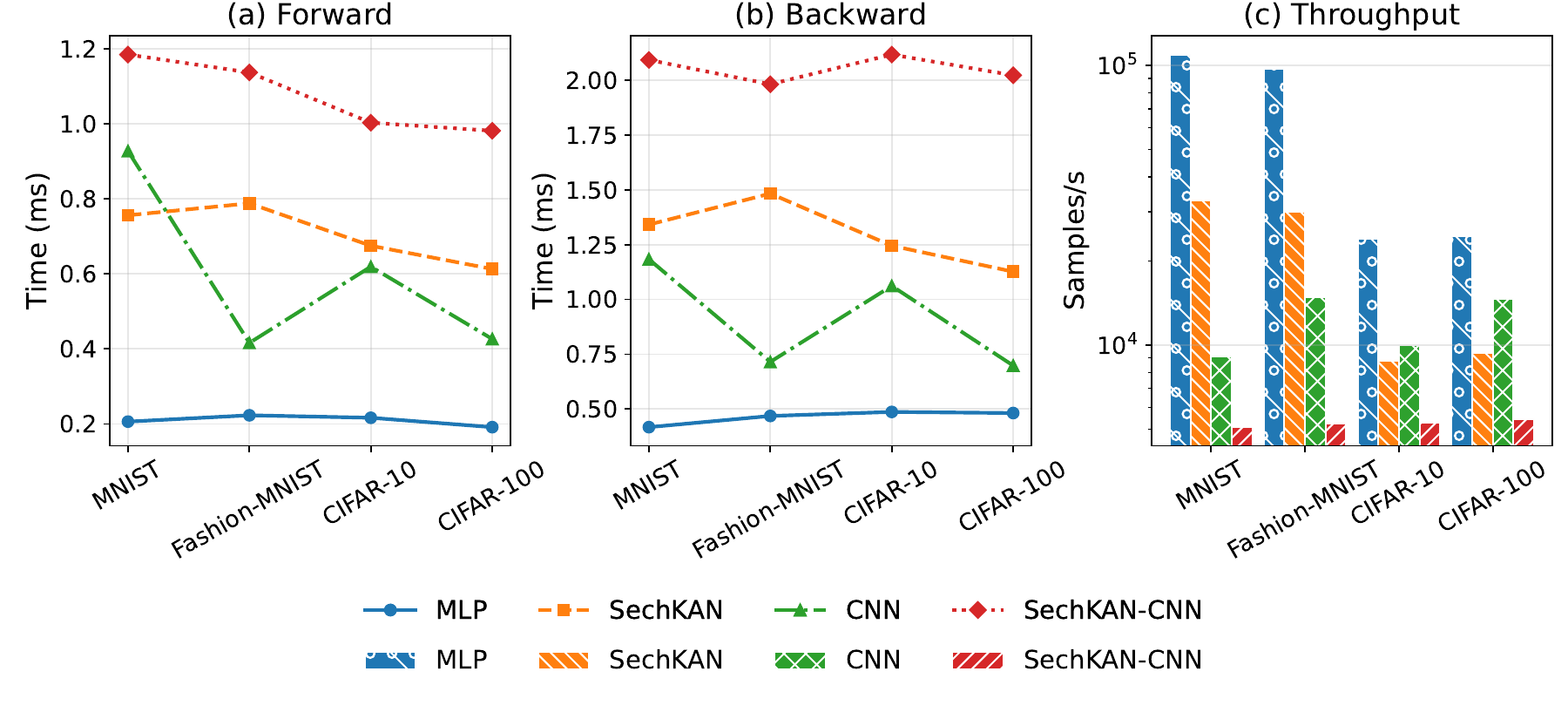}
  \centering
  \caption{Forward pass time, backward pass time, and training throughput of MLP, SechKAN, CNN, and SechKAN-CNN across different datasets.}
\label{fig:ic_forward_backward_throughput}
\end{figure*}

For MLP and SechKAN, we use the architecture \texttt{(784, 64, 10)} with a batch size of 64 for MNIST and Fashion-MNIST. For CIFAR-10 and CIFAR-100, input images are flattened to a dimension of 3072, and the architectures \texttt{(3072, 64, 10)} and \texttt{(3072, 256, 100)} are used respectively, with a batch size of 16 to control computational cost. In SechKAN, \texttt{num\_grids=4} is used for MNIST and Fashion-MNIST, while \texttt{num\_grids=8} is used for CIFAR datasets. We set \texttt{norm1\_type=None} (Norm1) and \texttt{norm2\_type=layer} (Norm2) for all SechKAN models. SechKAN also omits the skip connection (\texttt{use\_base\_update=False}) and disables the learnable width parameter (\texttt{use\_width=False}).

For CNN-based models, both CNN and SechKAN-CNN use the same structure:
Input → Conv1 → 2D Pool → Conv2 → 2D Pool → Flatten → MLP/SechKAN classification head → Output.
The shared convolutional backbone consists of two convolutional layers with \texttt{in\_channel}, \texttt{middle\_channel}, and \texttt{out\_channel}, followed by non-linear activations and 2D max pooling layers. The channel sizes are typically set to 1/8/16 for MNIST and Fashion-MNIST, and 3/24/48 for CIFAR datasets to handle grayscale and RGB inputs. After feature extraction, the output is flattened into a vector of size \texttt{flatten\_dim}. This vector is passed to a two-layer classifier with a hidden layer of size \texttt{hidden\_size} and an output layer of size \texttt{num\_classes}. CNN uses a standard MLP classifier with fully connected layers. In contrast, SechKAN-CNN replaces this head with SechKAN layers using sech basis functions and grid projections, where \texttt{num\_grids=4} is applied in layers for all datasets.

\begin{figure*}[htbp]
  \centering
\includegraphics[scale=0.5]{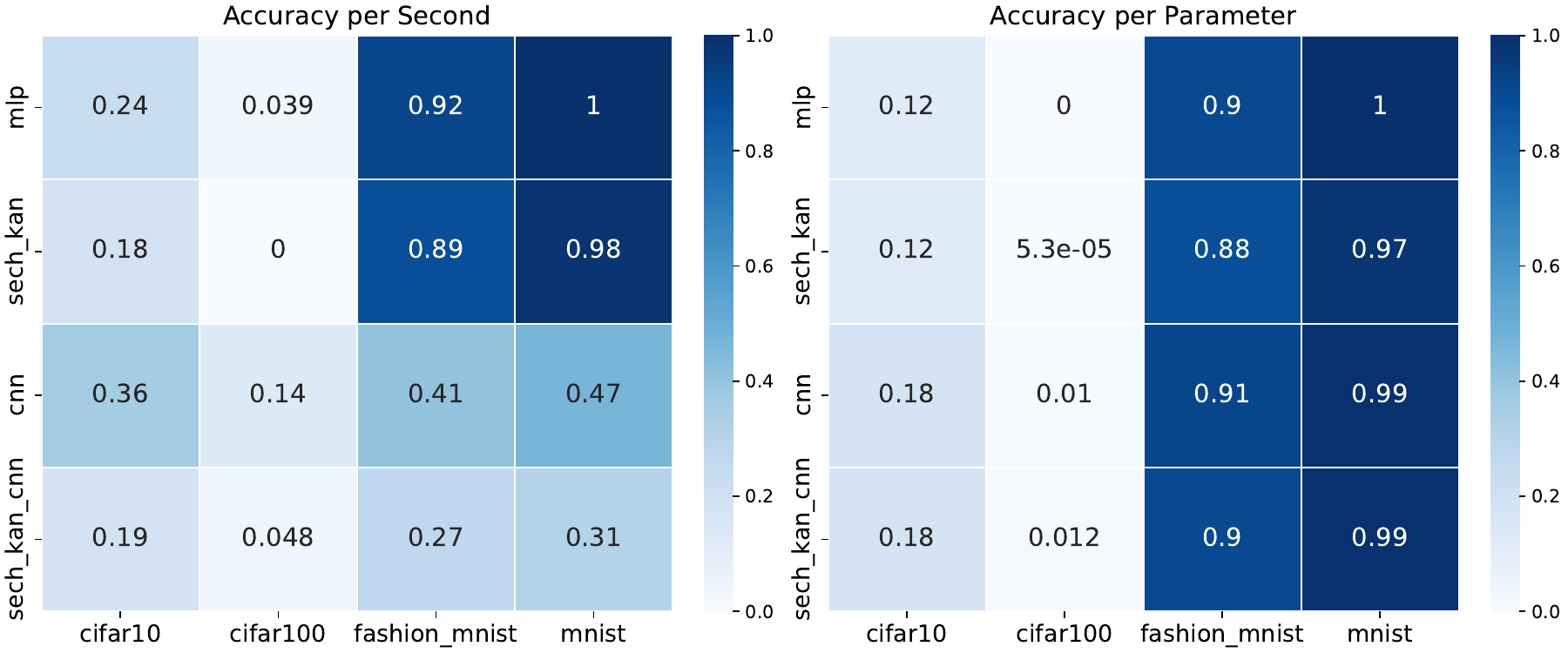}
  \centering
  \caption{Validation accuracy per second and per parameter, normalized to [0, 1], reflecting training efficiency. A value of 1 indicates the best result, while 0 indicates the worst.}
\label{fig:ic_training_eff}
\end{figure*}

The results in \Cref{tab:ic_result} highlight a trade-off between predictive performance and computational efficiency. In all datasets, SechKAN achieves higher validation accuracy and macro F1 than the MLP, although the magnitude of the improvement varies substantially. The gains are marginal on MNIST and CIFAR-100 but more noticeable on Fashion-MNIST and CIFAR-10, suggesting that the benefit of the sech-based representation is task-dependent. In the CNN-based setting, standard CNN performs slightly better on MNIST and Fashion-MNIST, whereas SechKAN-CNN achieves higher validation accuracy and macro F1 on CIFAR-10 and CIFAR-100, resulting in the highest average validation accuracy over datasets. Notably, on the CIFAR datasets, SechKAN-CNN obtains higher validation performance despite lower training accuracy than the standard CNN, suggesting improved generalization and a reduced tendency toward overfitting. These gains, however, come at the cost of increased computational time, with SechKAN-CNN requiring the longest training time on average.

The performance gains of SechKAN are accompanied by increased computational cost. As shown in \Cref{tab:ic_model_params}, SechKAN consistently requires higher MFLOPs and peak GPU memory than MLP while having only slightly higher parameter counts. For example, on MNIST, the computational cost increases from 0.051 to 0.096 MFLOPs, whereas on CIFAR-100, it increases from 0.813 to 1.135 MFLOPs. Similarly, replacing the MLP classifier in CNN with SechKAN results in only marginal changes in the number of parameters but consistently increases both MFLOPs and peak memory usage. This overhead is also reflected in \Cref{fig:ic_forward_backward_throughput}, where SechKAN exhibits longer forward and backward execution times than MLP, while SechKAN-CNN consistently incurs the longest execution times and lowest training throughput across all datasets.

To further assess the balance between predictive performance and computational cost, \Cref{fig:ic_training_eff} summarizes training efficiency using normalized validation accuracy per second and validation accuracy per parameter, with detailed values provided in \Cref{tab:ic_training_eff_appendix}. SechKAN remains competitive on MNIST and Fashion-MNIST but becomes less efficient on CIFAR-10 and particularly CIFAR-100. Despite its higher execution cost, SechKAN-CNN achieves parameter efficiency comparable to CNN, suggesting a more favorable balance between representational capability and training efficiency than SechKAN alone.

\section{Ablation Study}
In this section, we conduct ablation studies to analyze the behavior of SechKAN. We first examine the individual contributions of the sech basis function and the 1D projection, followed by the effects of the number of grids, activation functions, as well as the choice and placement of normalization methods. The configurations used in the main experiments (\Cref{sec:ex_eva}) serve as the baseline settings. For each study, only the component of interest is varied while all other settings remain fixed. Each configuration is evaluated using two or three random seeds (0, 1, and 2, depending on the experiment), and the reported results are averaged over the corresponding runs.

\subsection{Understanding SechKAN's Performance}

To better understand the source of SechKAN's performance improvements, we conduct an ablation study that separately evaluates the contributions of its two key components: the sech basis function and the 1D projection. The study considers two basis functions (B-spline and sech) and two architectural settings (with and without the 1D projection), yielding four comparable models: EfficientKAN, which employs a B-spline basis without the 1D projection; Modified EfficientKAN, which augments the B-spline basis with the 1D projection; Modified SechKAN, which uses the sech basis while omitting the 1D projection; and SechKAN, which combines the sech basis with the 1D projection.

For a fair comparison, all normalization layers are removed, the four models have approximately the same number of trainable parameters, and all other training settings remain identical. The ablation study is conducted on the function fitting and image classification benchmarks. PDE surrogate modeling tasks are excluded because they rely on early stopping, causing different models to converge after different numbers of epochs and follow different optimization trajectories.

As shown in~\Cref{tab:ablation_basis_projection_ff}, the B-spline basis generally achieves lower approximation error than the sech basis, whereas the sech basis consistently requires less training time. Furthermore, the effectiveness of the 1D projection depends on the choice of basis function. When combined with the B-spline basis, the 1D projection does not improve approximation accuracy or runtime and instead leads to a small degradation in both metrics. In contrast, when paired with the proposed sech basis, it noticeably improves approximation accuracy while incurring only a small increase in training time.


\begin{table}[!t]
\centering
\caption{Ablation study of the sech basis function and 1D projection on the function-fitting benchmarks. Results are averaged over 10 functions and 3 random seeds (0, 1, and 2). Lower loss and runtime indicate better performance. All models are configured to have approximately the same number of trainable parameters, and no normalization layers are used.}
\label{tab:ablation_basis_projection_ff}
\begin{tabular}{llllll}
\hline
\textbf{Model} & \textbf{Basis} & \textbf{1D Projection} & \textbf{Loss}$\downarrow$ & \textbf{Params} & \textbf{Runtime (s)}$\downarrow$ \\
\hline
EfficientKAN          & B-spline     & $\times$ &
$\mathbf{2.03\times10^{-2}\pm5.48\times10^{-2}}$ &
648 &
7.02$\pm$2.43 \\

Modified EfficientKAN & B-spline      & $\checkmark$ &
$2.76\times10^{-2}\pm7.55\times10^{-2}$ &
647 &
7.28$\pm$2.52 \\

Modified SechKAN      & sech  & $\times$ &
$5.26\times10^{-2}\pm8.24\times10^{-2}$ &
653 &
\textbf{6.01$\pm$1.84} \\

SechKAN (Proposed)    & sech  & $\checkmark$ &
$3.94\times10^{-2}\pm7.60\times10^{-2}$ &
660 &
6.22$\pm$1.85 \\
\hline
\end{tabular}
\end{table}

\begin{table*}[!t]
\centering
\caption{Ablation study of the sech and B-spline basis functions and the 1D projection on image classification datasets. Results are averaged over 3 random seeds (0, 1, and 2). Higher validation accuracy and lower training time indicate better performance. All models are configured to have approximately the same number of trainable parameters.}
\label{tab:ablation_basis_projection_cls}
\begin{tabular}{llllll}
\hline
\textbf{Dataset} & \textbf{Model} & \textbf{Basis} & \textbf{1D Projection} & \textbf{Val. Acc.}$\uparrow$ & \textbf{Runtime (s)}$\downarrow$ \\
\hline

\multirow{4}{*}{MNIST}
& EfficientKAN          & B-spline & $\times$     & $92.09\pm0.22$ & $86.37\pm1.52$ \\
& Modified EfficientKAN & B-spline & $\checkmark$ & $96.91\pm0.06$ & $86.15\pm2.65$ \\
& Modified SechKAN      & sech     & $\times$     & $95.52\pm0.09$ & $\mathbf{71.08\pm2.26}$ \\
& SechKAN (Proposed)    & sech     & $\checkmark$ & $\mathbf{97.12\pm0.21}$ & $71.69\pm3.84$ \\
\hline

\multirow{4}{*}{Fashion-MNIST}
& EfficientKAN          & B-spline & $\times$     & $85.94\pm0.06$ & $88.21\pm1.89$ \\
& Modified EfficientKAN & B-spline & $\checkmark$ & $88.01\pm0.12$ & $86.25\pm4.05$ \\
& Modified SechKAN      & sech     & $\times$     & $87.30\pm0.08$ & $70.03\pm4.94$ \\
& SechKAN (Proposed)    & sech     & $\checkmark$ & $\mathbf{88.36\pm0.18}$ & $\mathbf{66.74\pm2.74}$ \\
\hline

\multirow{4}{*}{CIFAR-10}
& EfficientKAN          & B-spline & $\times$     & $43.52\pm0.11$ & $172.75\pm14.98$ \\
& Modified EfficientKAN & B-spline & $\checkmark$ & $50.91\pm0.46$ & $166.59\pm2.67$ \\
& Modified SechKAN      & sech     & $\times$     & $39.83\pm2.16$ & $\mathbf{118.84\pm5.48}$ \\
& SechKAN (Proposed)    & sech     & $\checkmark$ & $\mathbf{51.07\pm0.15}$ & $133.51\pm5.00$ \\
\hline

\multirow{4}{*}{CIFAR-100}
& EfficientKAN          & B-spline & $\times$     & $22.32\pm0.19$ & $170.83\pm13.30$ \\
& Modified EfficientKAN & B-spline & $\checkmark$ & $\mathbf{25.87\pm0.08}$ & $162.98\pm0.83$ \\
& Modified SechKAN      & sech     & $\times$     & $8.01\pm6.91$ & $\mathbf{128.86\pm2.28}$ \\
& SechKAN (Proposed)    & sech     & $\checkmark$ & $24.96\pm0.54$ & $137.97\pm3.49$ \\
\hline
\end{tabular}
\end{table*}

A similar trend is observed for image classification (\Cref{tab:ablation_basis_projection_cls}). Consistent with the function-fitting results, the proposed sech basis substantially reduces training time across all datasets. On the simpler MNIST and Fashion-MNIST datasets, removing the 1D projection causes only a small drop in validation accuracy, while the sech basis continues to outperform the B-spline basis with comparable or lower training time. In contrast, on the more challenging CIFAR-10 and CIFAR-100 datasets, the sech basis alone performs worse than the B-spline basis, indicating that it is more difficult to optimize for complex visual recognition tasks. However, when combined with the 1D projection, SechKAN achieves the highest validation accuracy on CIFAR-10, whereas Modified SechKAN becomes unstable on CIFAR-100 because optimization encounters NaN gradients and diverges. These results indicate that the 1D projection stabilizes optimization and enables the sech basis to realize its full potential on more challenging tasks.

\subsection{Grid sizes}

\begin{figure*}[!t]
  \centering
\includegraphics[scale=0.52]{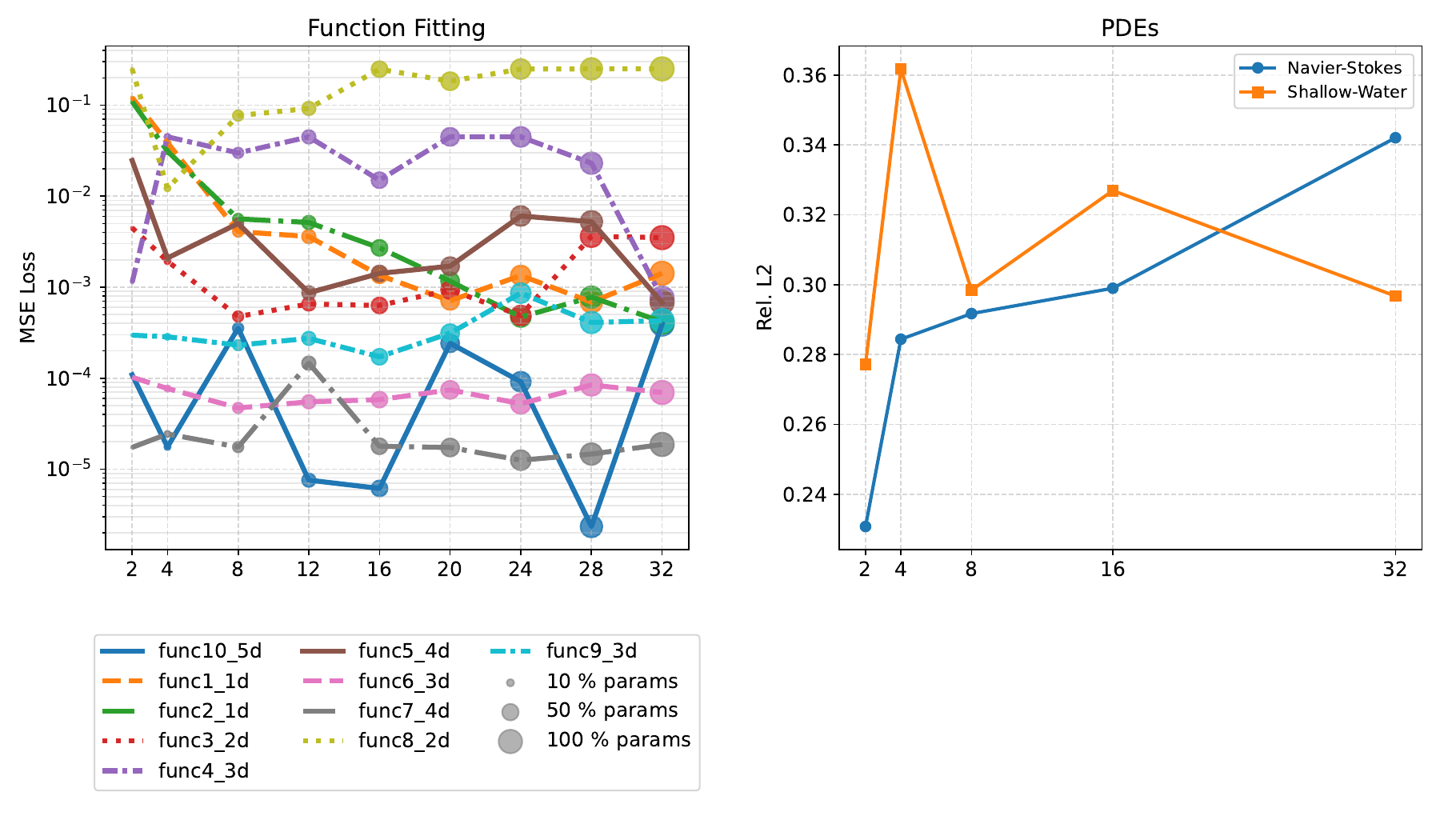}
  \centering
\caption{SechKAN results with different numbers of grids, averaged over two seeds (0 and 1). (Left) MSE loss on 10 function-fitting tasks ($f_1$--$f_{10}$)} with grid sizes from 2 to 32. (Right) Relative L2 error on the Navier--Stokes and Shallow Water datasets with grid sizes of 2, 4, 8, 16, and 32. Lower values (MSE loss or relative L2 error) indicate better performance.
\label{fig:ab_ff_pde_grid}
\end{figure*}

\begin{figure*}[!t]
  \centering
\includegraphics[scale=0.63]{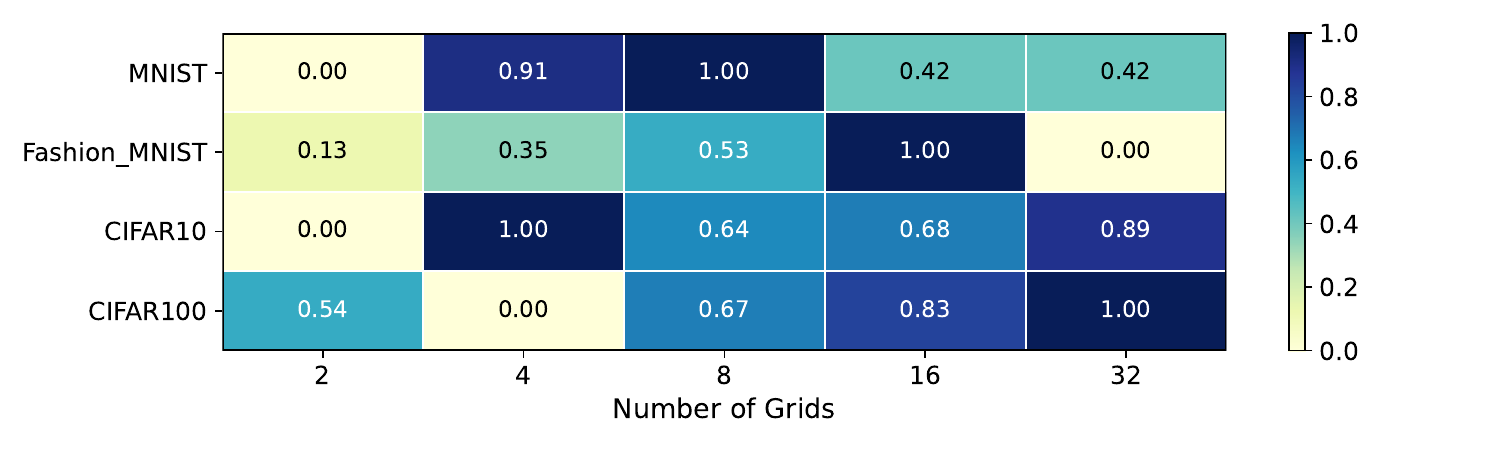}
  \centering
\caption{Heatmap of SechKAN validation accuracies, scaled from 0 to 1, with different numbers of grids on image classification datasets, averaged over two seeds (0 and 1). A value of 1 indicates the best result, while 0 indicates the worst.}
\label{fig:ab_ic_grid}
\end{figure*}


We evaluate SechKAN with grid sizes of 2, 4, 8, 12, 16, 20, 24, 28, and 32 for function-fitting tasks and grid sizes of 2, 4, 8, 16, and 32 for the Navier--Stokes and Shallow Water datasets. As shown in \Cref{fig:ab_ff_pde_grid}, larger grid sizes (20--32) often achieve the lowest MSE on most functions, including functions $f_1$--$f_5$, $f_7$, and $f_{10}$, although they also increase the number of trainable parameters. For the PDE datasets, smaller grid sizes are generally competitive, while the optimal choice depends on the specific problem. For example, a grid size of 4 performs worse than several larger configurations on the Shallow Water dataset.

As shown in \Cref{fig:ab_ic_grid}, in image classification, grid sizes of 8 and 16 are often strong choices across datasets, while grid sizes of 4 and 32 can achieve peak performance in specific cases (e.g., grid size 4 on CIFAR-10 and grid size 32 on CIFAR-100). However, no single grid size is consistently optimal, and grid size 2 is generally the weakest performer, indicating limited sensitivity to grid size in this setting. Larger grids increase model capacity but do not provide clear and consistent gains. Thus, no single grid size performs best across all tasks, as performance is task-dependent despite increased capacity from larger grids.


\subsection{Activation Functions}

\begin{figure*}[!t]
  \centering
\includegraphics[scale=0.72]{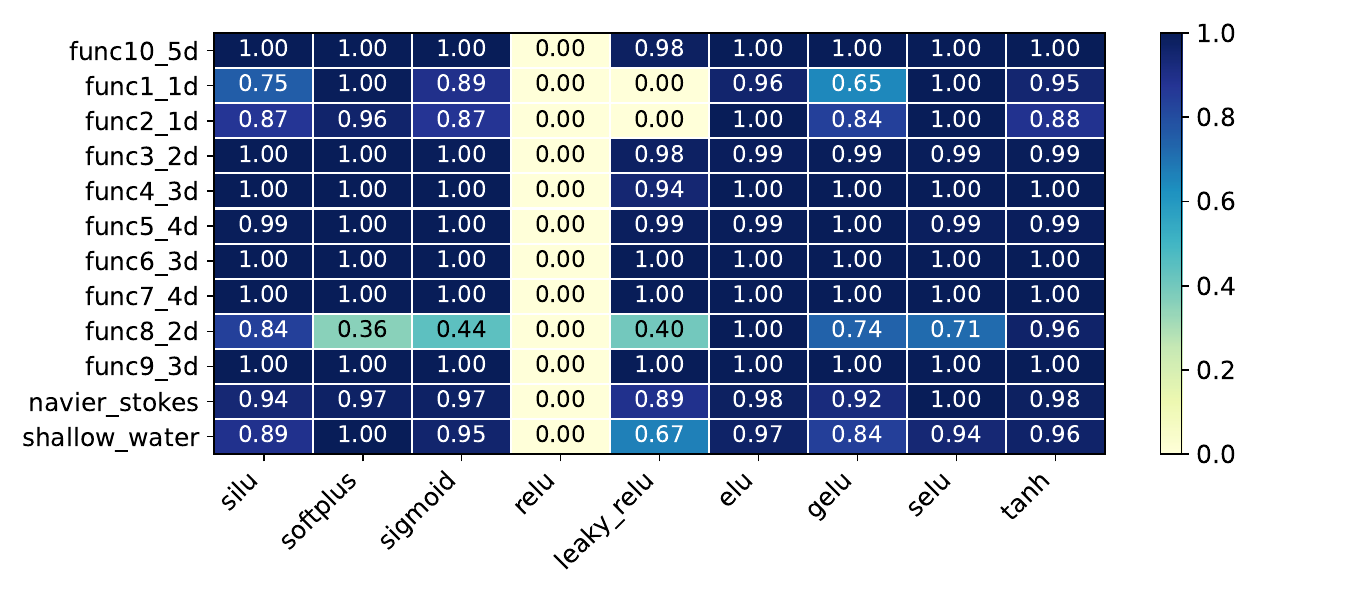}
  \centering
\caption{Heatmap of SechKAN results with different activation functions, averaged over two seeds (0 and 1). MSE loss and relative L2 error are normalized to $[0,1]$ for each function or PDE problem (Navier--Stokes and Shallow Water). A value of 1 indicates the best result, while 0 indicates the worst.}
\label{fig:ab_ff_pde_act}
\end{figure*}

\begin{figure*}[!t]
  \centering
\includegraphics[scale=0.63]{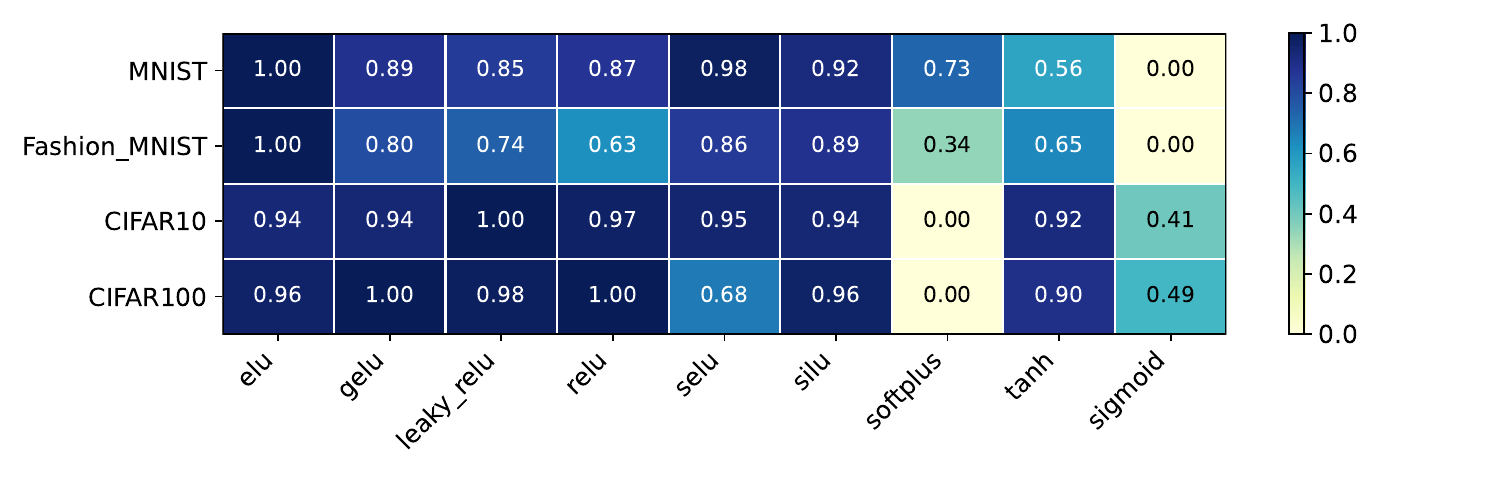}
  \centering
\caption{Heatmap of SechKAN validation accuracies with different activation functions on image classification datasets, averaged over two seeds (0 and 1). Validation accuracies are normalized to $[0,1]$. A value of 1 indicates the best result, while 0 indicates the worst.}
\label{fig:ab_ic_act}
\end{figure*}

\Cref{fig:ab_ff_pde_act} presents the performance of SechKAN on function fitting and PDE surrogate modeling using different activation functions. ELU, SELU, and tanh consistently achieve the best performance across most tasks, followed closely by SiLU and GELU. Softplus and sigmoid also produce competitive results on the majority of benchmarks, although both perform noticeably worse on $f_8$. In contrast, Leaky ReLU exhibits weak performance on $f_1$ and $f_2$, while ReLU consistently performs the worst across all function-fitting and PDE tasks.

As shown in \Cref{fig:ab_ic_act}, most activation functions work well with SechKAN on image classification datasets, with the notable exceptions of Softplus and Sigmoid, which perform poorly across all datasets. Among the well-performing activations, ELU and GELU achieve the strongest results, closely followed by ReLU, SELU, and SiLU (with Leaky ReLU also performing competitively on several datasets). Although activation choice affects validation accuracy, the performance gap among strong activations is relatively small, indicating that SechKAN is largely robust to the choice of activation function.


\subsection{Data Normalization}

\begin{figure*}[!t]
  \centering
\includegraphics[scale=0.72]{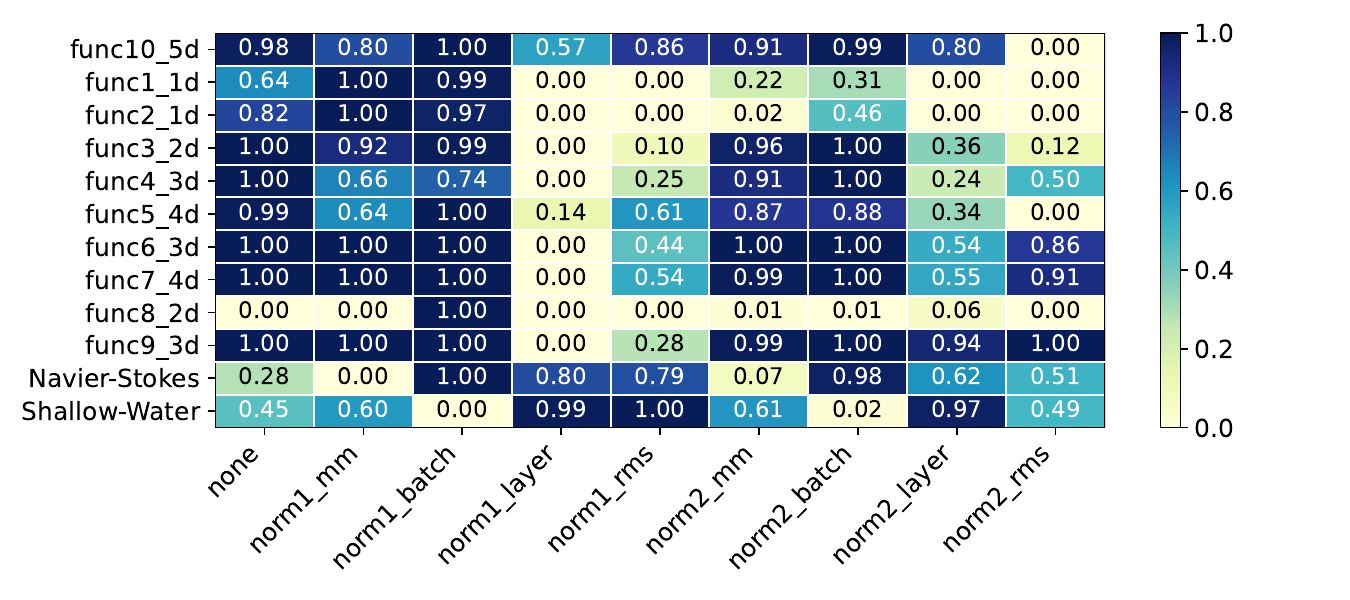}
  \centering
\caption{Heatmap of SechKAN results averaged over two seeds (0 and 1) under different data normalization methods: no normalization, Norm1 or Norm2 positions with Min--Max, LayerNorm, BatchNorm, and RMSNorm. MSE loss and relative L2 error are normalized to $[0,1]$ for each function or PDE problem (Navier--Stokes and Shallow Water). A value of 1 indicates the best result, while 0 indicates the worst.}
\label{fig:ab_ff_pde_norm}
\end{figure*}

\begin{figure*}[!t]
  \centering
\includegraphics[scale=0.63]{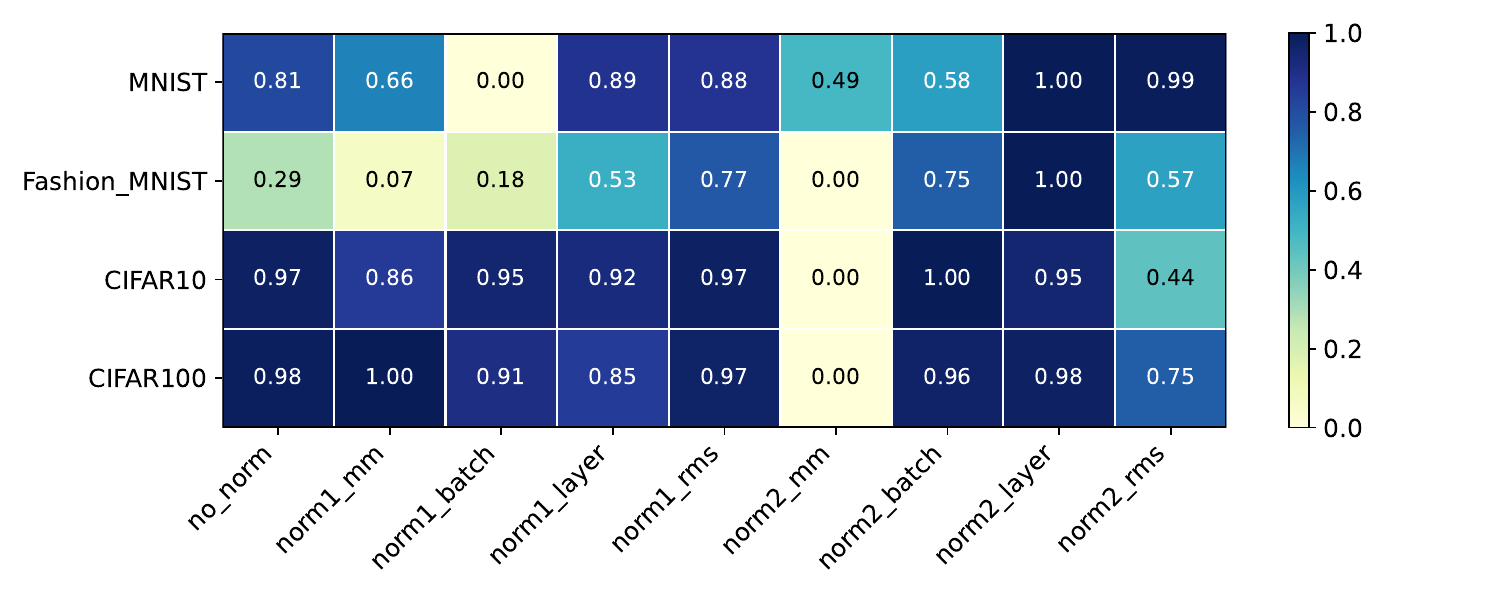}
  \centering
\caption{Heatmap of SechKAN validation accuracies on image classification datasets averaged over two seeds (0 and 1) for different data normalization methods: no normalization, and Min--Max, LayerNorm, BatchNorm, or RMSNorm applied at the Norm1 or Norm2 position. A value of 1 indicates the best result, whereas 0 indicates the worst result.}
\label{fig:ab_ic_norm}
\end{figure*}

\Cref{fig:ab_ff_pde_norm} shows the performance of SechKAN on function fitting and PDE surrogate modeling using different normalization strategies. To reduce the number of parameters, only a single normalization layer is applied, either before the sech basis transformation (Norm1) or after the grid projection (Norm2).

For function fitting, nearly all normalization configurations, as well as the model without normalization, fail on $f_8$, with the only exception being BatchNorm in Norm1. BatchNorm in Norm1 achieves the best performance across the benchmark functions, followed by Min--Max normalization in Norm1 and the model without normalization. In contrast, BatchNorm and Min--Max normalization in Norm2 fail on several functions, particularly $f_1$, $f_2$, and $f_8$. LayerNorm and RMSNorm generally perform poorly on function-fitting tasks regardless of whether they are applied in Norm1 or Norm2.

For PDE surrogate modeling, LayerNorm and RMSNorm in Norm1 produce the strongest overall performance on both the Navier--Stokes and Shallow Water datasets. BatchNorm is effective only for the Navier--Stokes dataset. LayerNorm in Norm2 also performs competitively, particularly for the Shallow Water dataset, whereas most of the remaining normalization configurations yield inferior performance or fail to converge.

\Cref{fig:ab_ic_norm} shows the results of SechKAN with different data normalization methods on image classification datasets. In short, LayerNorm applied at Norm2 achieves the best accuracy across datasets, followed by RMSNorm at Norm1 and LayerNorm at Norm1. Some normalization methods perform well on only one or two datasets but fail to generalize. For example, Min--Max normalization at Norm1 works well on CIFAR-10 and CIFAR-100 but performs poorly when applied at Norm2. Fashion-MNIST is particularly sensitive to the choice of normalization and only achieves strong performance when LayerNorm is applied at Norm2.

We find that Min--Max normalization, BatchNorm, and, in some cases, no normalization generally yield the best performance for function-fitting tasks, whereas LayerNorm and RMSNorm are more suitable for PDE surrogate modeling and image classification. Furthermore, the placement of the normalization layer has a substantial impact on performance, with Norm1 and Norm2 often favoring different normalization methods. These results suggest that normalization should be treated as a task-dependent design choice rather than a universal component of SechKAN.

\section{Discussion}

Despite promising results across several tasks, SechKAN has several limitations. First, its performance has not been evaluated on a broader set of AI Feynman benchmark functions or on large-scale datasets such as ImageNet. Second, there is no systematic method for selecting the number of grids or normalization strategies, so these choices rely on heuristic tuning. Third, SechKAN requires slightly longer training times than MLPs for image classification, indicating room for improving training efficiency. These limitations can be addressed by evaluating SechKAN on more diverse benchmarks, developing automated methods for grid and normalization selection, and improving computational efficiency through architectural refinements and optimized training algorithms.

Based on the ablation studies, SechKAN shows limited-to-moderate sensitivity to grid size, whereas its sensitivity to normalization strategy and placement is more pronounced. Increasing the grid size generally improves representational capacity but does not consistently yield better performance across tasks, indicating that no single grid size is universally optimal. We therefore recommend selecting the grid size through validation on the target dataset. In our experiments, moderate grid sizes (e.g., 4–8) provided a reasonable starting point for many tasks.

The optimal normalization strategy is task-dependent. For function fitting, BatchNorm before basis expansion (Norm1) consistently achieves the best performance, while Min--Max normalization and, in some cases, no normalization also perform competitively. For PDE surrogate modeling, LayerNorm at Norm1 provides the strongest overall performance, although BatchNorm is also effective for the Navier–Stokes dataset and LayerNorm at Norm2 remains competitive for the Shallow Water dataset. For image classification, LayerNorm at Norm2 consistently achieves the best performance across datasets. These observations are further supported by the gradient-flow analysis in \Cref{sec:sech_kan_gradient_flow}, which shows that appropriate normalization improves optimization by mitigating activation saturation and maintaining stable gradient propagation, although its effectiveness depends on both the normalization method and its placement. Therefore, rather than recommending a universal normalization strategy, we suggest selecting both the normalization method and its placement according to the target task and validating the choice empirically.

\section{Conclusion}

In this paper, we proposed SechKAN, a novel KAN that employs hyperbolic secant (sech) functions as localized basis functions. Unlike many existing KAN variants, SechKAN combines a lightweight 1D linear projection with the sech basis expansion, allowing the model to maintain a number of trainable parameters comparable to that of MLPs while preserving the nonlinear representation capability of KANs. The proposed architecture also supports optional normalization modules, skip connections, and learnable scale, bias, and width parameters, providing a flexible framework for different learning settings.

Extensive experiments on function fitting, PDE surrogate modeling, and image classification show that SechKAN achieves competitive or superior predictive performance compared with representative KAN variants while maintaining parameter efficiency comparable to that of MLPs. In particular, SechKAN consistently outperformed the evaluated KAN variants on image classification tasks and remained competitive on function fitting and PDE surrogate modeling. Beyond predictive performance, we conducted comprehensive analyses of parameter count, FLOPs, GPU memory usage, execution time, training throughput, and gradient flow to characterize the computational properties of SechKAN. In addition, the ablation studies quantify the individual contributions of the sech basis function and the 1D projection, while providing practical insights into the effects of grid size, activation functions, and normalization strategies across different learning tasks.

Although SechKAN substantially reduces the parameter count of conventional KANs, it still incurs higher computational cost than standard MLPs because of basis-function evaluation. Moreover, no single architectural configuration was consistently optimal across all tasks, indicating that task-specific choices of grid size, normalization, and activation function remain important. Future work will focus on improving computational efficiency, investigating adaptive and learnable basis functions, exploring more expressive parameter-efficient KANs, and extending SechKAN to larger-scale learning problems and additional scientific and engineering applications.

\newpage
\appendix
\section{Comparison of Sech and Other Basis Functions in KANs}
\label{sec:sech_analysis}

\begin{figure*}[!t]
\centering
\includegraphics[scale=0.54]{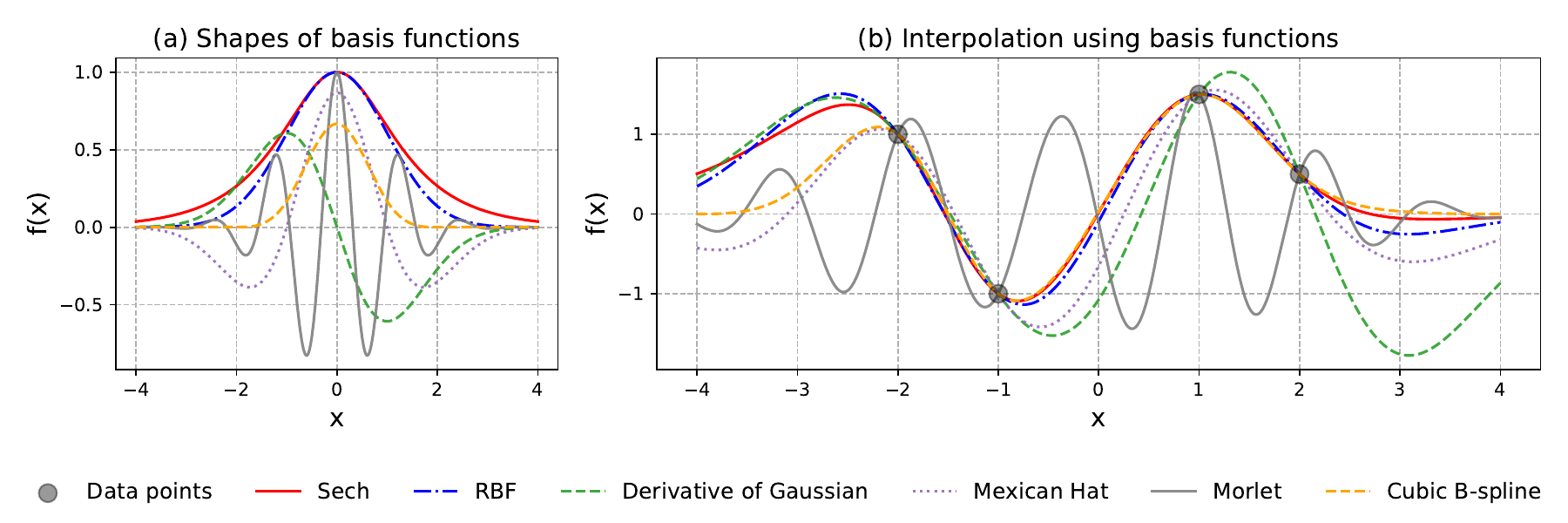}
\caption{Comparison of basis functions: (a) basis function shapes; (b) interpolation of four data points using each basis function.}
\label{fig:function_plot}
\end{figure*}

In this section, we compare the proposed sech basis with several representative basis functions commonly used in KANs~\cite{noorizadegan2026practitioner}. \Cref{tab:function_formula} summarizes their mathematical formulations, while \Cref{fig:function_plot} illustrates their shapes and interpolation behavior. Gaussian RBF, sech, and cubic B-spline produce bell-shaped curves and smooth interpolations of the given data points, whereas the derivative of Gaussian and the wavelet-based functions exhibit oscillatory behavior, with Morlet showing the strongest oscillations. Among these basis functions, sech combines exponential decay and infinite differentiability, supporting stable gradient propagation during optimization. In contrast, Gaussian RBF is more localized, the derivative of Gaussian introduces sign changes, and cubic B-splines rely on piecewise definitions that increase implementation complexity.

\begin{table}[!t]
\centering
\caption{Some popular basis functions used in KANs.}
\begin{tabular}{ll}
\hline
\textbf{Function} & \textbf{Formula} \\
\hline
sech &
$\displaystyle \mathrm{sech}(x)=\frac{1}{\cosh(x)}$ \\
\hline
Gaussian Radial Basis Function (Gaussian RBF) &
$\displaystyle \psi(x)=e^{-x^2/2}$ \\
\hline
Derivative of Gaussian &
$\displaystyle \psi(x)=-x\,e^{-x^2/2}$ \\
\hline
Mexican hat wavelet &
$\displaystyle \psi(x)=\frac{2}{\sqrt{3}\pi^{1/4}}(1-x^2)e^{-x^2/2}$ \\
\hline
Morlet wavelet &
$\displaystyle \psi(x)=\cos(\omega x)e^{-x^2/2}, \quad \omega=5.0$ \\
\hline
Cubic B-spline &
$\displaystyle
B(x)=
\begin{cases}
\displaystyle\frac{\mkern4mu 4-6|x|^2+3|x|^3\mkern4mu}{6}, & |x|<1,\\[1.8ex]
\displaystyle\frac{(2-|x|)^3}{6}, & 1\le|x|<2,\\
0, & \text{otherwise}.
\end{cases}
$ \\
\hline
\end{tabular}
\label{tab:function_formula}
\end{table}

\begin{table}[!t]
\centering
\caption{Evaluation runtime of basis functions implemented using native PyTorch CUDA operators. An input tensor of size $1000$ is evaluated for each basis function. Results are averaged over $1{,}000{,}000$ runs (mean $\pm$ standard deviation in milliseconds).}
\begin{tabular}{lc}
\hline
\textbf{Basis Function} & \textbf{Runtime (ms) $\downarrow$} \\
\hline
Sech & $\mathbf{0.040 \pm 0.127}$ \\
Gaussian RBF & $0.050 \pm 0.203$ \\
Derivative of Gaussian & $0.051 \pm 0.212$ \\
Mexican hat wavelet & $0.146 \pm 0.599$ \\
Morlet wavelet ($\omega=5$) & $0.059 \pm 0.156$ \\
Cubic B-spline & $0.229 \pm 0.449$ \\
\hline
\end{tabular}
\label{tab:function_runtime}
\end{table}

We first compare the computational cost of evaluating the basis functions independently of any KAN implementation. Each function is implemented using native PyTorch CUDA operators and evaluated on the same input tensor. As shown in \Cref{tab:function_runtime}, the sech basis achieves the lowest evaluation runtime, indicating that its mathematical formulation is well suited for efficient GPU execution.

\begin{table}[!t]
\centering
\caption{Forward-pass runtime of representative KAN basis implementations in PyTorch (CUDA). All implementations use eight basis functions and are evaluated on an input tensor of size $(64,784)$. Runtime is reported as the mean $\pm$ standard deviation (ms) over $20{,}000$ forward passes after $100$ warm-up iterations.}
\label{tab:basis_runtime}
\begin{tabular}{lc}
\hline
\textbf{Basis Function} & \textbf{Runtime (ms) $\downarrow$} \\
\hline
B-spline (EfficientKAN) & $0.367 \pm 0.017$ \\
Bare sech (SechKAN) & $\mathbf{0.031 \pm 0.002}$ \\
Parametric sech (SechKAN) & $0.081 \pm 0.002$ \\
Gaussian RBF (FastKAN) & $0.056 \pm 0.004$ \\
RSWAF (FasterKAN) & $0.052 \pm 0.004$ \\
\hline
\end{tabular}
\end{table}

To evaluate whether this computational advantage translates to practical KANs, we further benchmark representative basis implementations from EfficientKAN (cubic B-spline), FastKAN (Gaussian RBF), FasterKAN (RSWAF), and SechKAN (bare sech and parametric sech) under identical PyTorch/CUDA settings. All implementations employ eight basis functions (corresponding to the number of grids for Sech/RBF networks and the grid size plus spline order for B-spline networks like EfficientKAN) and are evaluated on a CUDA-enabled GPU using an input tensor of size $(64,784)$. Runtime is measured over $20{,}000$ forward passes after $100$ warm-up iterations.

Bare sech is the parameter-free basis function, whereas parametric sech introduces learnable width, scale, and bias parameters. As shown in \Cref{tab:basis_runtime}, the bare sech basis achieves the lowest forward-pass runtime, requiring approximately $12.5\times$ less execution time than the representative cubic B-spline implementation while remaining competitive with other element-wise basis functions. This improvement is primarily due to its implementation using only dense element-wise tensor operations, whereas the cubic B-spline implementation relies on interval selection and recursive Cox--de Boor basis evaluation. Introducing learnable width, scale, and bias parameters increases the runtime of the parametric sech variant to $0.081$ ms, but it remains approximately $4.5\times$ faster than the representative B-spline implementation. This comparison is limited to representative implementations under the same PyTorch/CUDA environment and does not claim universal superiority over all spline-based methods.

\section{Experimental Details for Function Fitting}
\label{sec:ff_appendix}

\begin{table*}[!ht]
\small
\centering
\caption{Performance comparison of different models on functions $f_1$--$f_10$ (mean $\pm$ standard deviation over 10 seeds).}
\begin{tabular}{p{2.8cm}p{2cm}p{2.2cm}p{4cm}p{2.5cm}}
\hline
\textbf{Model} & \textbf{Function} & \textbf{Used Params} & \textbf{Loss (MSE)} $\downarrow$ & \textbf{Runtime (s)} $\downarrow$ \\
\hline

BSRBF-KAN & $f_1$ & 22 & $3.33\times10^{-2} \pm 2.06\times10^{-3}$ & $2.91 \pm 0.11$ \\
FastKAN & $f_1$ & 22 & $1.00\times10^{-1} \pm 1.77\times10^{-3}$ & $2.70 \pm 0.09$ \\
ReLU-KAN & $f_1$ & 22 & $6.76\times10^{-2} \pm 1.22\times10^{-2}$ & $2.80 \pm 0.12$ \\
SechKAN & $f_1$ & 22 & $4.32\times10^{-3} \pm 3.72\times10^{-3}$ & $3.29 \pm 0.15$ \\
EfficientKAN & $f_1$ & 22 & $\mathbf{2.37\times10^{-3} \pm 8.43\times10^{-6}}$ & $2.97 \pm 0.05$ \\
\hline

BSRBF-KAN & $f_2$ & 22 & $4.74\times10^{-2} \pm 1.04\times10^{-3}$ & $2.92 \pm 0.10$ \\
FastKAN & $f_2$ & 22 & $6.20\times10^{-2} \pm 2.37\times10^{-3}$ & $2.71 \pm 0.08$ \\
ReLU-KAN & $f_2$ & 22 & $4.11\times10^{-2} \pm 6.95\times10^{-3}$ & $2.81 \pm 0.07$ \\
SechKAN & $f_2$ & 22 & $\mathbf{8.88\times10^{-3} \pm 6.41\times10^{-3}}$ & $3.30 \pm 0.05$ \\
EfficientKAN & $f_2$ & 22 & $1.19\times10^{-2} \pm 1.60\times10^{-3}$ & $2.95 \pm 0.14$ \\
\hline

BSRBF-KAN & $f_3$ & 364 & $4.59\times10^{-4} \pm 3.15\times10^{-4}$ & $8.65 \pm 0.39$ \\
FastKAN & $f_3$ & 373 & $6.17\times10^{-4} \pm 4.97\times10^{-4}$ & $7.18 \pm 0.17$ \\
ReLU-KAN & $f_3$ & 345 & $8.46\times10^{-4} \pm 7.34\times10^{-4}$ & $7.36 \pm 0.16$ \\
SechKAN & $f_3$ & 363 & $1.48\times10^{-3} \pm 1.81\times10^{-3}$ & $8.43 \pm 0.09$ \\
EfficientKAN & $f_3$ & 364 & $\mathbf{2.71\times10^{-4} \pm 2.71\times10^{-4}}$ & $8.47 \pm 0.42$ \\
\hline

BSRBF-KAN & $f_4$ & 448 & $\mathbf{7.92\times10^{-4} \pm 2.71\times10^{-4}}$ & $7.21 \pm 0.23$ \\
FastKAN & $f_4$ & 457 & $2.69\times10^{-3} \pm 2.46\times10^{-3}$ & $5.58 \pm 0.14$ \\
ReLU-KAN & $f_4$ & 441 & $5.78\times10^{-3} \pm 1.30\times10^{-2}$ & $5.69 \pm 0.10$ \\
SechKAN & $f_4$ & 459 & $2.27\times10^{-2} \pm 2.32\times10^{-2}$ & $6.65 \pm 0.20$ \\
EfficientKAN & $f_4$ & 448 & $1.59\times10^{-3} \pm 6.34\times10^{-4}$ & $6.86 \pm 0.34$ \\
\hline

BSRBF-KAN & $f_5$ & 1456 & $8.38\times10^{-3} \pm 2.63\times10^{-3}$ & $8.78 \pm 0.35$ \\
FastKAN & $f_5$ & 1473 & $7.55\times10^{-3} \pm 4.20\times10^{-3}$ & $7.24 \pm 0.18$ \\
ReLU-KAN & $f_5$ & 1457 & $2.13\times10^{-2} \pm 5.44\times10^{-3}$ & $7.27 \pm 0.14$ \\
SechKAN & $f_5$ & 1445 & $\mathbf{1.89\times10^{-3} \pm 1.91\times10^{-3}}$ & $8.54 \pm 0.16$ \\
EfficientKAN & $f_5$ & 1456 & $2.78\times10^{-3} \pm 1.17\times10^{-3}$ & $8.51 \pm 0.40$ \\
\hline
BSRBF-KAN & $f_6$ & 448 & $4.47\times10^{-5} \pm 1.38\times10^{-5}$ & $6.17 \pm 0.35$ \\
FastKAN & $f_6$ & 457 & $2.97\times10^{-5} \pm 1.16\times10^{-5}$ & $4.66 \pm 0.11$ \\
ReLU-KAN & $f_6$ & 441 & $6.30\times10^{-5} \pm 1.87\times10^{-5}$ & $4.78 \pm 0.10$ \\
SechKAN & $f_6$ & 459 & $1.02\times10^{-4} \pm 1.03\times10^{-4}$ & $5.70 \pm 0.12$ \\
EfficientKAN & $f_6$ & 448 & $\mathbf{1.23\times10^{-5} \pm 5.05\times10^{-6}}$ & $5.68 \pm 0.31$ \\
\hline

BSRBF-KAN & $f_7$ & 1456 & $2.04\times10^{-5} \pm 1.12\times10^{-5}$ & $7.93 \pm 0.40$ \\
FastKAN & $f_7$ & 1473 & $2.21\times10^{-5} \pm 9.13\times10^{-6}$ & $6.43 \pm 0.13$ \\
ReLU-KAN & $f_7$ & 1457 & $1.60\times10^{-5} \pm 7.85\times10^{-6}$ & $6.46 \pm 0.08$ \\
SechKAN & $f_7$ & 1445 & $1.92\times10^{-5} \pm 1.52\times10^{-5}$ & $7.68 \pm 0.14$ \\
EfficientKAN & $f_7$ & 1456 & $\mathbf{2.50\times10^{-6} \pm 1.25\times10^{-6}}$ & $7.75 \pm 0.39$ \\
\hline

BSRBF-KAN & $f_8$ & 364 & $2.26\times10^{-1} \pm 2.82\times10^{-2}$ & $8.80 \pm 0.34$ \\
FastKAN & $f_8$ & 373 & $2.50\times10^{-1} \pm 4.22\times10^{-4}$ & $7.17 \pm 0.13$ \\
ReLU-KAN & $f_8$ & 345 & $2.37\times10^{-1} \pm 2.35\times10^{-2}$ & $7.23 \pm 0.14$ \\
SechKAN & $f_8$ & 363 & $\mathbf{1.32\times10^{-1} \pm 9.42\times10^{-2}}$ & $8.36 \pm 0.17$ \\
EfficientKAN & $f_8$ & 364 & $1.71\times10^{-1} \pm 2.19\times10^{-2}$ & $8.50 \pm 0.34$ \\
\hline

BSRBF-KAN & $f_9$ & 448 & $2.95\times10^{-5} \pm 1.64\times10^{-5}$ & $5.76 \pm 0.25$ \\
FastKAN & $f_9$ & 457 & $3.05\times10^{-5} \pm 2.18\times10^{-5}$ & $4.58 \pm 0.17$ \\
ReLU-KAN & $f_9$ & 441 & $6.47\times10^{-5} \pm 2.39\times10^{-5}$ & $4.66 \pm 0.13$ \\
SechKAN & $f_9$ & 459 & $3.06\times10^{-3} \pm 9.05\times10^{-3}$ & $5.60 \pm 0.10$ \\
EfficientKAN & $f_9$ & 448 & $\mathbf{3.65\times10^{-6} \pm 1.41\times10^{-6}}$ & $5.59 \pm 0.34$ \\
\hline

BSRBF-KAN & $f_{10}$ & 1456 & $5.96\times10^{-3} \pm 1.81\times10^{-3}$ & $9.81 \pm 0.45$ \\
FastKAN & $f_{10}$ & 1473 & $9.64\times10^{-3} \pm 2.52\times10^{-3}$ & $8.21 \pm 0.10$ \\
ReLU-KAN & $f_{10}$ & 1467 & $1.14\times10^{-2} \pm 4.98\times10^{-3}$ & $8.29 \pm 0.06$ \\
SechKAN & $f_{10}$ & 1479 & $\mathbf{2.28\times10^{-6} \pm 2.14\times10^{-6}}$ & $9.82 \pm 0.18$ \\
EfficientKAN & $f_{10}$ & 1456 & $2.97\times10^{-3} \pm 5.93\times10^{-4}$ & $9.59 \pm 0.42$ \\
\hline

\end{tabular}
\label{tab:ff_detail}
\end{table*}

Experiments are repeated using random seeds $\{0,1,\ldots,9\}$. One-dimensional functions are sampled using uniformly spaced points over $[0,1]$, whereas higher-dimensional functions are evaluated on uniform Cartesian grids over $[0,1]^d$. Results are reported as the mean and standard deviation over the 10 runs.

The appendix reports the exact hyperparameter configurations used for each KAN variant. FastKAN uses grid sizes $\{20,20,12,13,13,13,13,12,13,12\}$ for $f_1$--$f_{10}$, respectively. BSRBF-KAN adopts spline order 3 with grid sizes $\{18,18,9,10,10,10,10,9,10,9\}$, while EfficientKAN uses spline order 3 with grid sizes $\{17,17,8,9,9,9,9,8,9,8\}$. ReLU-KAN uses spline order 3 with grid sizes $\{4,4,4,5,7,5,7,4,5,5\}$. For SechKAN, Min--Max normalization is used for $f_1$ and $f_2$, whereas BatchNorm is applied for $f_3$--$f_{10}$. The corresponding grid sizes are $\{8,8,4,4,8,4,8,4,4,8\}$.

\Cref{tab:ff_detail} presents the complete results for all functions, including the number of trainable parameters, MSE, and training runtime. All models are compared under similar parameter budgets, while their approximation performance varies across functions of different complexity. SechKAN demonstrates competitive performance relative to the other KAN variants, achieving the lowest MSE on $f_2$, $f_5$, $f_8$, and $f_{10}$ while maintaining a comparable number of trainable parameters.

\section{Experimental Details for PDE surrogate modeling}
\label{sec:app_ex_detail_pde}



\begin{table}[ht]
\centering
\caption{Performance comparison of SechKAN variants on the Navier--Stokes dataset (mean $\pm$ std over 5 seeds).}
\begin{tabular}{p{3.2cm}p{2.7cm}p{2.7cm}p{2.7cm}p{2.2cm}}
\hline
\textbf{Model} &
\textbf{Val. Rel. L2 $\downarrow$} &
\textbf{Test Rel. L2 $\downarrow$} &
\textbf{Runtime (s) $\downarrow$} &
\textbf{Used Params} \\
\hline

SechKAN1 &
0.14179 $\pm$ 0.01390 &
0.31438 $\pm$ 0.02998 &
290.13 $\pm$ 2.10 &
38267 \\

SechKAN2 &
0.22064 $\pm$ 0.21106 &
0.39155 $\pm$ 0.20445 &
298.70 $\pm$ 9.58 &
39035 \\

SechKAN3 &
0.20465 $\pm$ 0.04908 &
0.33635 $\pm$ 0.04548 &
316.47 $\pm$ 67.48 &
38267 \\

SechKAN4 &
0.15595 $\pm$ 0.03044 &
0.31355 $\pm$ 0.03014 &
229.53 $\pm$ 50.08 &
38243 \\

\textbf{SechKAN5 (\Cref{tab:navier_stoke_comparison})} &
\textbf{0.13664 $\pm$ 0.01529} &
\textbf{0.27695 $\pm$ 0.00954} &
266.04 $\pm$ 1.87 &
39011 \\

SechKAN6 &
0.33738 $\pm$ 0.18304 &
0.41670 $\pm$ 0.13447 &
\textbf{197.55 $\pm$ 64.63} &
38243 \\

\hline
Average &
0.19951 &
0.34158 &
266.40 &
38511 \\

\hline
\end{tabular}
\label{tab:sechkan_variant_navier_stoke}
\end{table}

\begin{table}[h]
\centering
\caption{Performance comparison of SechKAN variants on the Shallow Water dataset (mean $\pm$ std over 5 seeds).}
\begin{tabular}{p{3.2cm}p{2.7cm}p{2.7cm}p{2.7cm}p{2.2cm}}
\hline
\textbf{Model} &
\textbf{Val. Rel. L2 $\downarrow$} &
\textbf{Test Rel. L2 $\downarrow$} &
\textbf{Runtime (s) $\downarrow$} &
\textbf{Used Params} \\
\hline

\textbf{SechKAN1 (\Cref{tab:shallow_water_comparison})} &
\textbf{0.12637 $\pm$ 0.00567} &
\textbf{0.29879 $\pm$ 0.01425} &
196.59 $\pm$ 18.88 &
39228 \\

SechKAN2 &
0.21322 $\pm$ 0.10831 &
0.34455 $\pm$ 0.09023 &
209.61 $\pm$ 40.14 &
39204 \\

SechKAN3 &
0.29362 $\pm$ 0.12529 &
0.39399 $\pm$ 0.10176 &
\textbf{87.34 $\pm$ 19.05} &
38436 \\

SechKAN4 &
0.18429 $\pm$ 0.11008 &
0.33093 $\pm$ 0.10245 &
167.14 $\pm$ 66.28 &
38436 \\

\hline
Average &
0.20438 &
0.34207 &
165.17 &
38826 \\
\hline
\end{tabular}
\label{tab:sechkan_variant_shallow_water}
\end{table}

The Navier--Stokes dataset is generated using a two-dimensional pseudo-spectral solver on a periodic domain $[0,2\pi]\times[0,2\pi]$ with a $64\times64$ spatial grid. The temporal evolution is computed using a fourth-order Runge--Kutta (RK4) scheme with a time step of $\Delta t=10^{-2}$ for 60 time steps. The viscosity coefficient is set to $\nu=10^{-3}$, and an external forcing term is applied during the simulation. The initial vorticity field is randomly sampled from a Gaussian distribution and filtered in the Fourier domain using a $2/3$-rule dealiasing scheme. The velocity components $(u,v)$ are recovered from the vorticity field through the stream-function formulation and used as output physical quantities. The Shallow Water dataset is generated using a two-dimensional pseudo-spectral solver on a structured $64\times64$ spatial grid over the domain $[0,1]\times[0,1]$. The equations are integrated using RK4 with a time step of $\Delta t=5\times10^{-4}$ for 60 time steps, with $g=9.81$ and $\nu=2\times10^{-3}$. The bathymetry is fixed as a Gaussian--sinusoidal field, while the initial height and velocity fields are analytically prescribed. The solution is evolved in conservative variables $(h,hu,hv)$, and the resulting $(h,u,v)$ fields are used as targets for spatio-temporal inputs $(x,y,t)$.

We evaluate multiple variants with different numbers of grids to balance model capacity and training efficiency, as well as different normalization types and positions. We also experiment with various normalization types, but LayerNorm performs best and is used in the main experiments. For the normalization mode, we compare applying no normalization, applying it to all layers, and applying it to all layers except the first, and adopt the latter in the final configuration due to the test results. We set \texttt{use\_width = False} and \texttt{use\_base\_update = False} to better isolate the contribution of the hyperbolic secant basis functions.

For the Navier--Stokes dataset, we evaluate six SechKAN variants. All variants share the same backbone architecture $[in\_dim, 192, 192, out\_dim]$, SiLU base activation, \texttt{use\_width = False}, and \texttt{use\_base\_update = False}, with normalization applied to all hidden layers except the input layer (\texttt{norm\_mode = "except\_first"}). SechKAN1, SechKAN2, and SechKAN3 use \texttt{num\_grids = 8}, whereas SechKAN4, SechKAN5, and SechKAN6 use \texttt{num\_grids = 4}. SechKAN1 and SechKAN4 do not employ normalization (\texttt{norm1\_type = ""}, \texttt{norm2\_type = ""}). SechKAN2 and SechKAN5 apply LayerNorm at Norm1 (\texttt{norm1\_type = "layer"}), whereas SechKAN3 and SechKAN6 apply Min--Max normalization at Norm2 (\texttt{norm2\_type = "mm"}).

For the Shallow Water dataset, we evaluate four SechKAN variants with the same backbone architecture and training configuration. SechKAN1 employs LayerNorm at Norm1 (\texttt{norm1\_type = "layer"}) and uses \texttt{num\_grids = 8}. SechKAN2 uses the same normalization setting but with \texttt{num\_grids=4}. SechKAN3 applies Min--Max normalization at Norm2 (\texttt{norm2\_type = "mm"}) with \texttt{num\_grids = 4}, whereas SechKAN4 does not use normalization (\texttt{norm1\_type = ""}, \texttt{norm2\_type = ""}) and also uses \texttt{num\_grids = 4}.

Based on the results in \Cref{tab:sechkan_variant_navier_stoke}, SechKAN5 is selected for the main experiments reported in \Cref{tab:navier_stoke_comparison} on the Navier--Stokes dataset. For the Shallow Water dataset, SechKAN1 achieves the lowest test Relative L2 error in \Cref{tab:sechkan_variant_shallow_water} and is therefore selected for the experiments reported in \Cref{tab:shallow_water_comparison}.

\section{Additional Experiments for Image Classification}
\label{sec:ic_appendix}

In \Cref{tab:ic_training_eff_appendix}, we evaluate the training efficiency of SechKAN and SechKAN-CNN in comparison with MLP and CNN. Efficiency is measured as validation accuracy normalized by training time (seconds) and by the number of parameters. Parameter counts are reported in \Cref{tab:ic_model_params}, while validation accuracy and training time are taken from \Cref{tab:ic_result}. MLP achieves the highest average accuracy-per-second, with SechKAN close behind. CNN achieves the highest average accuracy-per-parameter, followed closely by SechKAN-CNN. On MNIST and Fashion-MNIST, all four models have similar accuracy-per-parameter, whereas on CIFAR-10 and CIFAR-100, SechKAN-CNN achieves the highest accuracy-per-parameter, only slightly outperforming CNN. These results indicate that SechKAN is competitive with MLP in training efficiency, while SechKAN-CNN achieves accuracy-per-parameter comparable to CNN but incurs lower accuracy-per-second because of its higher computational cost.


\begin{table*}[!htbp]
\centering
\caption{Efficiency comparison (validation accuracy per second and per parameter) of MLP, SechKAN, CNN, and SechKAN-CNN across datasets.}
\begin{tabular}{p{2.5cm}p{2.5cm}p{3cm}p{3cm}}
\hline
\textbf{Dataset} & \textbf{Network} & \textbf{Acc / Time (s) $\uparrow$} & \textbf{Acc / Param $\uparrow$} \\
\hline

\multirow{4}{*}{MNIST}
& MLP & $\mathbf{1.40 \times 10^{-2}}$ & $\mathbf{1.91 \times 10^{-3}}$ \\
& SechKAN & $1.38 \times 10^{-2}$ & $1.86 \times 10^{-3}$ \\
& SechKAN-CNN & $5.50 \times 10^{-3}$ & $1.90 \times 10^{-3}$ \\
& CNN & $7.48 \times 10^{-3}$ & $1.90 \times 10^{-3}$ \\
\hline

\multirow{4}{*}{Fashion-MNIST}
& MLP & $\mathbf{1.30 \times 10^{-2}}$ & $1.73 \times 10^{-3}$ \\
& SechKAN & $1.26 \times 10^{-2}$ & $1.69 \times 10^{-3}$ \\
& SechKAN-CNN & $5.05 \times 10^{-3}$ & $1.73 \times 10^{-3}$ \\
& CNN & $6.72 \times 10^{-3}$ & $\mathbf{1.75 \times 10^{-3}}$ \\
\hline

\multirow{4}{*}{CIFAR-10}
& MLP & $4.62 \times 10^{-3}$ & $2.62 \times 10^{-4}$ \\
& SechKAN & $3.98 \times 10^{-3}$ & $2.58 \times 10^{-4}$ \\
& SechKAN-CNN & $4.01 \times 10^{-3}$ & $\mathbf{3.65 \times 10^{-4}}$ \\
& CNN & $\mathbf{6.07 \times 10^{-3}}$ & $3.61 \times 10^{-4}$ \\
\hline

\multirow{4}{*}{CIFAR-100}
& MLP & $2.20 \times 10^{-3}$ & $2.95 \times 10^{-5}$ \\
& SechKAN & $1.72 \times 10^{-3}$ & $2.95 \times 10^{-5}$ \\
& SechKAN-CNN & $2.31 \times 10^{-3}$ & $\mathbf{5.21 \times 10^{-5}}$ \\
& CNN & $\mathbf{3.39 \times 10^{-3}}$ & $4.92 \times 10^{-5}$ \\
\hline

\multirow{4}{*}{Average}
& MLP & $\mathbf{8.46 \times 10^{-3}}$ & $9.83 \times 10^{-4}$ \\
& SechKAN & $8.03 \times 10^{-3}$ & $9.59 \times 10^{-4}$ \\
& SechKAN-CNN & $4.22 \times 10^{-3}$ & $1.01 \times 10^{-3}$ \\
& CNN & $5.92 \times 10^{-3}$ & $\mathbf{1.02 \times 10^{-3}}$ \\
\hline
\end{tabular}
\label{tab:ic_training_eff_appendix}
\end{table*}

\begin{table*}[!ht]
\centering
\caption{Model configurations, parameter counts, MFLOPs, and peak GPU memory of different KAN variants in image classification experiments.}
\begin{tabular}{p{2.3cm}p{2.3cm}p{3cm}p{1.5cm}p{1.8cm}p{1.7cm}}
\hline
\textbf{Dataset} & \textbf{Model} & \textbf{Network structure} & \textbf{Used Params} & \textbf{MFLOPs $\downarrow$} & \textbf{Peak Mem. (MB) $\downarrow$}\\
\hline

\multirow{5}{*}{\textbf{\shortstack[l]{MNIST\\Fashion-MNIST}}}
& SechKAN       & (784, 64, 10) & 52,608 & 0.096 & \textbf{23.84} \\
& BSRBF-KAN     & (784,7,10) & 51,604 & 0.324 & 41.55 \\
& FastKAN       & (784,7,10) & 51,621 & 0.108 & 26.59 \\
& FasterKAN     & (784,8,10) & 52,400 & \textbf{0.086} & 26.64 \\
& EfficientKAN  & (784,7,10) & 50,022 & 0.273 & 24.36 \\
\hline

\multirow{5}{*}{\textbf{CIFAR-10}}
& SechKAN       &  (3072, 64, 10) & 203,632 & 0.501 & 31.84 \\
& BSRBF-KAN     & (3072,7,10) & 200,324 & 1.260 & 43.12 \\
& FastKAN       & (3072,7,10) & 200,341 & 0.419 & 28.75 \\
& FasterKAN     & (3072,8,10) & 203,408 & \textbf{0.336} & 28.75 \\
& EfficientKAN  & (3072,7,10) & 215,740 & 1.195 & \textbf{27.11} \\
\hline

\multirow{5}{*}{\textbf{CIFAR-100}}
& SechKAN       & (3072, 256, 100) & 819,082 & 1.135 & 43.58 \\
& BSRBF-KAN     & (3072,28,100) & 805,544 & 1.872 & 51.80 \\
& FastKAN       & (3072,28,100) & 805,672 & 1.026 & 38.61 \\
& FasterKAN     & (3072,32,100) & 818,240 & \textbf{0.952} & 38.14 \\
& EfficientKAN  & (3072,28,100) & 799,344 & 2.076 & \textbf{34.39} \\
\hline

\end{tabular}
\label{tab:kan_model_complexity}
\end{table*}

\begin{table*}[!ht]
\centering
\caption{Performance comparison of KAN variants across datasets. Results are averaged over 5 runs, with one run per seed (0, 1, 2, 3, and 4).}
\begin{tabular}{p{2.5cm}p{2.5cm}p{2.2cm}p{2.1cm}p{2.1cm}p{2cm}}
\hline
\textbf{Dataset} & \textbf{Network} & \textbf{Train. Acc. $\uparrow$} & \textbf{Val. Acc. $\uparrow$} & \textbf{Val. F1 $\uparrow$} & \textbf{Runtime (s)} $\downarrow$\\
\hline

\multirow{6}{*}{MNIST}
& BSRBF-KAN    & $94.47 \pm 0.13$ & $92.54 \pm 0.08$ & $92.38 \pm 0.08$ & $84.22 \pm 2.32$ \\
& EfficientKAN & $92.88 \pm 0.19$ & $92.27 \pm 0.17$ & $92.14 \pm 0.18$ & $86.26 \pm 1.85$ \\
& FastKAN      & $94.79 \pm 0.12$ & $93.23 \pm 0.14$ & $93.08 \pm 0.14$ & $71.69 \pm 2.24$ \\
& FasterKAN    & $93.06 \pm 0.09$ & $92.70 \pm 0.08$ & $92.53 \pm 0.09$ & $\mathbf{67.86 \pm 2.26}$ \\
& MLP          & $97.97 \pm 0.02$ & $97.28 \pm 0.05$ & $97.24 \pm 0.05$ & $69.59 \pm 5.22$ \\
& SechKAN      & $\mathbf{99.52 \pm 0.07}$ & $\mathbf{97.68 \pm 0.08}$ & $\mathbf{97.65 \pm 0.08}$ & $70.73 \pm 2.32$ \\
\hline

\multirow{6}{*}{Fashion-MNIST}
& BSRBF-KAN    & $91.60 \pm 0.07$ & $87.10 \pm 0.09$ & $87.05 \pm 0.08$ & $85.40 \pm 2.19$ \\
& EfficientKAN & $88.40 \pm 0.07$ & $85.92 \pm 0.03$ & $85.85 \pm 0.04$ & $84.12 \pm 5.32$ \\
& FastKAN      & $91.09 \pm 0.08$ & $87.29 \pm 0.06$ & $87.23 \pm 0.07$ & $70.66 \pm 1.23$ \\
& FasterKAN    & $88.88 \pm 0.08$ & $86.46 \pm 0.14$ & $86.31 \pm 0.14$ & $\mathbf{67.64 \pm 2.77}$ \\
& MLP          & $90.57 \pm 0.01$ & $87.93 \pm 0.04$ & $87.85 \pm 0.04$ & $67.69 \pm 5.63$ \\
& SechKAN      & $\mathbf{92.82 \pm 0.06}$ & $\mathbf{89.03 \pm 0.07}$ & $\mathbf{88.95 \pm 0.07}$ & $70.69 \pm 4.25$ \\
\hline

\multirow{6}{*}{CIFAR-10}
& BSRBF-KAN    & $56.97 \pm 0.42$ & $42.64 \pm 0.20$ & $42.33 \pm 0.20$ & $191.75 \pm 6.33$ \\
& EfficientKAN & $48.06 \pm 0.25$ & $43.38 \pm 0.12$ & $42.57 \pm 0.12$ & $163.86 \pm 6.06$ \\
& FastKAN      & $56.35 \pm 0.26$ & $43.77 \pm 0.24$ & $43.45 \pm 0.21$ & $155.00 \pm 2.37$ \\
& FasterKAN    & $47.90 \pm 0.20$ & $44.04 \pm 0.05$ & $43.63 \pm 0.08$ & $135.35 \pm 6.45$ \\
& MLP          & $61.18 \pm 0.15$ & $51.69 \pm 0.24$ & $51.35 \pm 0.24$ & $111.98 \pm 5.81$ \\
& SechKAN      & $\mathbf{64.46 \pm 0.91}$ & $\mathbf{52.59 \pm 0.15}$ & $\mathbf{52.38 \pm 0.15}$ & $\mathbf{132.23 \pm 8.70}$ \\
\hline

\multirow{6}{*}{CIFAR-100}
& BSRBF-KAN    & $46.96 \pm 0.38$ & $20.37 \pm 0.17$ & $19.42 \pm 0.17$ & $199.66 \pm 4.40$ \\
& EfficientKAN & $29.83 \pm 0.11$ & $22.43 \pm 0.11$ & $21.01 \pm 0.13$ & $162.13 \pm 2.63$ \\
& FastKAN      & $40.63 \pm 0.70$ & $21.82 \pm 0.16$ & $20.86 \pm 0.14$ & $131.32 \pm 3.96$ \\
& FasterKAN    & $29.10 \pm 0.07$ & $22.78 \pm 0.09$ & $21.06 \pm 0.08$ & $\mathbf{127.13 \pm 3.73}$ \\
& MLP          & $44.03 \pm 0.13$ & $23.94 \pm 0.12$ & $23.17 \pm 0.14$ & $108.71 \pm 9.00$ \\
& SechKAN      & $\mathbf{55.56 \pm 2.24}$ & $\mathbf{24.20 \pm 2.16}$ & $\mathbf{23.65 \pm 2.10}$ & $140.89 \pm 2.50$ \\
\hline

\multirow{5}{*}{Average}
& BSRBF-KAN    & $72.50$ & $60.66$ & $60.30$ & $140.26$ \\
& EfficientKAN & $64.79$ & $61.00$ & $60.39$ & $124.09$ \\
& FastKAN      & $70.72$ & $61.53$ & $61.16$ & $107.17$ \\
& FasterKAN    & $64.73$ & $61.49$ & $60.88$ & $\mathbf{99.50}$ \\
& MLP & 73.44 & 65.21 & 64.90 & 89.49 \\
& SechKAN      & $\mathbf{78.09}$ & $\mathbf{65.88}$ & $\mathbf{65.66}$ & $103.64$ \\
\hline

\end{tabular}
\label{tab:ic_extra_result}
\end{table*}

\begin{figure*}[!ht]
  \centering
\includegraphics[scale=0.52]{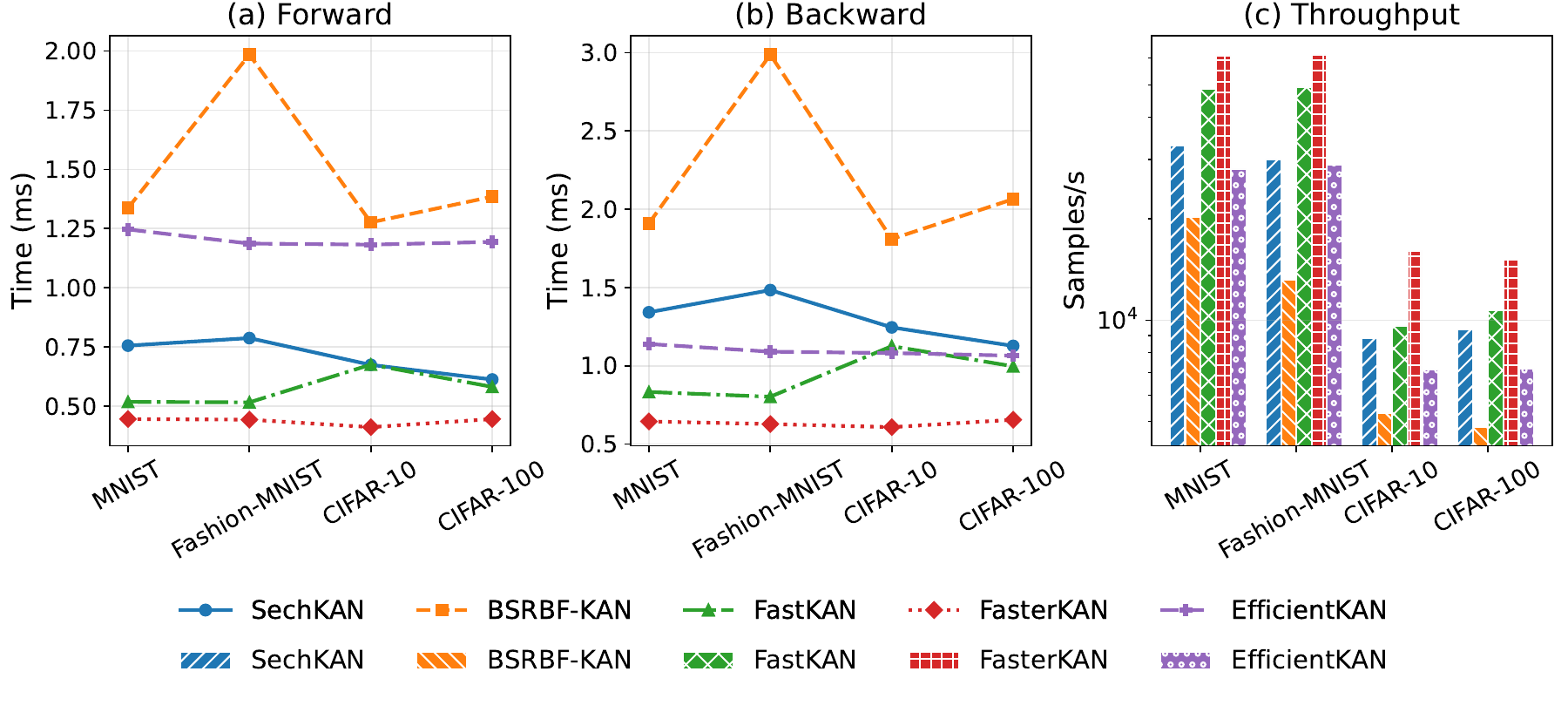}
  \centering
  \caption{Forward pass time, backward pass time, and training throughput of SechKAN, BSRBF-KAN, FastKAN, FasterKAN, and EfficientKAN across different datasets.}
\label{fig:ic_forward_backward_throughput_extended}
\end{figure*}

To further evaluate the effectiveness of SechKAN, we compare it with MLP and several representative KAN variants, including BSRBF-KAN, FastKAN, FasterKAN, EfficientKAN. For a fair comparison, each model is configured by adjusting its network architecture and hyperparameters (e.g., the number of grids and spline order) so that all models have approximately the same number of trainable parameters as the baseline MLP, as shown in \Cref{tab:kan_model_complexity}. 

As shown in \Cref{tab:ic_extra_result}, SechKAN achieves the highest training accuracy, validation accuracy, and validation macro F1 score on all datasets, outperforming the other KAN variants. The largest improvements are observed on MNIST and CIFAR-10, where SechKAN achieves clear gains in both validation accuracy and macro F1 score. Among the KAN variants, SechKAN is the only model that exceeds MLP in average validation accuracy and macro F1 score.

On average, SechKAN achieves the highest validation accuracy and macro F1 score among the compared KAN variants. Relative to FastKAN, which attains the second-highest average validation accuracy and macro F1 score, SechKAN improves these metrics by 4.35 and 4.50 percentage points, respectively. SechKAN has a parameter count comparable to the other KAN variants, with differences of less than 10\% across datasets. In terms of computational complexity, it requires fewer MFLOPs than BSRBF-KAN and EfficientKAN, whereas FastKAN and especially FasterKAN require fewer MFLOPs, with FasterKAN consistently exhibiting the lowest computational complexity. In terms of memory consumption, SechKAN achieves the lowest peak GPU memory usage on MNIST and Fashion-MNIST, whereas EfficientKAN is the most memory-efficient on CIFAR-10 and CIFAR-100, while FastKAN and FasterKAN exhibit memory requirements comparable to those of SechKAN.

The runtime measurements in \Cref{fig:ic_forward_backward_throughput_extended} are consistent with the average training times reported in \Cref{tab:ic_extra_result}. SechKAN generally requires shorter forward pass times than BSRBF-KAN and EfficientKAN, while its backward pass times are lower than those of BSRBF-KAN, FastKAN, and FasterKAN, remaining comparable to those of EfficientKAN. Consistent with these execution-time characteristics, SechKAN achieves higher training throughput than BSRBF-KAN and EfficientKAN, although it remains behind FastKAN and FasterKAN. FastKAN generally exhibits lower execution times than SechKAN, whereas FasterKAN consistently records the shortest forward and backward pass times, the highest training throughput, and the shortest average training time. Averaged across all datasets, SechKAN reduces training time by 26.1\%, 16.5\%, and 3.3\% compared with BSRBF-KAN, EfficientKAN, and FastKAN, respectively, while requiring 4.2\% more training time than FasterKAN and 15.8\% more than MLP.

\section{Analysis of Gradient Flow and Saturation in SechKAN}
\label{sec:sech_kan_gradient_flow}
 
To investigate the effects of normalization and the learnable basis width on gradient propagation, we trained 18 SechKAN variants on a synthetic regression task with the target function $y=\exp(-4x^{2})\sin(6\pi x)$, where $x$ is computed as the normalized sum of the input features. Each model consisted of six SechKAN layers with the architecture $[32,64,64,128,64,32,1]$. The synthetic dataset comprised 10{,}000 samples with 32-dimensional inputs. All models used the SiLU activation function and were trained with the AdamW optimizer (learning rate $10^{-3}$, weight decay $10^{-5}$) for 5{,}000 epochs.

The model configurations are summarized in \Cref{tab:sechkan_variants}. Two baseline models were trained without normalization, while the remaining variants employed LayerNorm, BatchNorm, RMSNorm, or Min--Max normalization at either the Norm1 position (before the sech basis transformation) or the Norm2 position (after the grid projection), with and without the learnable basis width. During training, the training loss and the L2 norm of the model gradients were recorded at every epoch to evaluate optimization behavior, gradient propagation, and the influence of normalization on activation saturation. Note that the results of this experiment should not be interpreted as representative of SechKAN's performance on all tasks.

\begin{table*}[!ht]
\centering
\caption{Configuration and empirical gradient statistics of the 18 SechKAN variants used for gradient flow analysis. Norm1 and Norm2 denote the two normalization positions illustrated in \Cref{fig:sechkan_archi}.}
\label{tab:sechkan_variants}
\begin{tabular}{llllllll}
\hline
\textbf{Model} &
\textbf{Norm1} &
\textbf{Norm2} &
\textbf{Width} &
\textbf{Best Loss} &
\textbf{Final Grad.} &
\textbf{Max Grad.} &
\textbf{Grad. Std.} \\
\hline
Model1  & None      & None      & No  & $1.61\times10^{-4}$ & $1.23\times10^{-1}$ & $1.92$ & $2.01\times10^{-1}$ \\
Model2  & None      & None      & Yes & $2.39\times10^{-5}$ & $1.82\times10^{-3}$ & $3.87$ & $3.44\times10^{-1}$ \\
\hline
Model3  & LayerNorm & None      & No  & $4.17\times10^{-7}$ & $4.54\times10^{-2}$ & $1.41$ & $1.44\times10^{-1}$ \\
Model4  & BatchNorm & None      & No  & $4.44\times10^{-8}$ & $1.24\times10^{-2}$ & $9.07\times10^{-1}$ & $3.88\times10^{-2}$ \\
Model5  & RMSNorm   & None      & No  & $1.57\times10^{-6}$ & $7.56\times10^{-2}$ & $1.33$ & $1.54\times10^{-1}$ \\
Model6  & Min--Max   & None      & No  & $9.56\times10^{-4}$ & $5.72\times10^{-2}$ & $11.05$ & $9.93\times10^{-1}$ \\
Model7  & LayerNorm & None      & Yes & $1.30\times10^{-8}$ & $4.27\times10^{-2}$ & $1.65$ & $1.03\times10^{-1}$ \\
Model8  & BatchNorm & None      & Yes & $6.26\times10^{-8}$ & $1.97\times10^{-3}$ & $6.23\times10^{-1}$ & $3.77\times10^{-2}$ \\
Model9  & RMSNorm   & None      & Yes & $1.03\times10^{-7}$ & $4.45\times10^{-5}$ & $2.25$ & $1.20\times10^{-1}$ \\
Model10 & Min--Max   & None      & Yes & $1.34\times10^{-4}$ & $5.50\times10^{-3}$ & $5.76$ & $3.97\times10^{-1}$ \\
\hline
Model11 & None      & LayerNorm & No  & $6.71\times10^{-8}$ & $6.58\times10^{-3}$ & $3.77\times10^{-1}$ & $2.14\times10^{-2}$ \\
Model12 & None      & BatchNorm & No  & $3.92\times10^{-8}$ & $2.01\times10^{-2}$ & $1.68$ & $3.91\times10^{-2}$ \\
Model13 & None      & RMSNorm   & No  & $8.88\times10^{-7}$ & $4.31\times10^{-2}$ & $5.67$ & $2.15\times10^{-1}$ \\
Model14 & None      & Min--Max   & No  & $1.18\times10^{-3}$ & $1.42\times10^{-1}$ & $1.51$ & $1.60\times10^{-1}$ \\
Model15 & None      & LayerNorm & Yes & $1.45\times10^{-8}$ & $2.05\times10^{-1}$ & $3.31$ & $1.53\times10^{-1}$ \\
Model16 & None      & BatchNorm & Yes & $4.17\times10^{-10}$ & $3.02\times10^{-2}$ & $1.84$ & $8.67\times10^{-2}$ \\
Model17 & None      & RMSNorm   & Yes & $8.00\times10^{-2}$ & $4.95\times10^{-8}$ & $6.51\times10^{-1}$ & $2.15\times10^{-2}$ \\
Model18 & None      & Min--Max   & Yes & $2.93\times10^{-6}$ & $2.80\times10^{-1}$ & $37.68$ & $6.33\times10^{-1}$ \\
\hline
\end{tabular}
\end{table*}

\begin{figure*}[!ht]
  \centering
\includegraphics[scale=0.32]{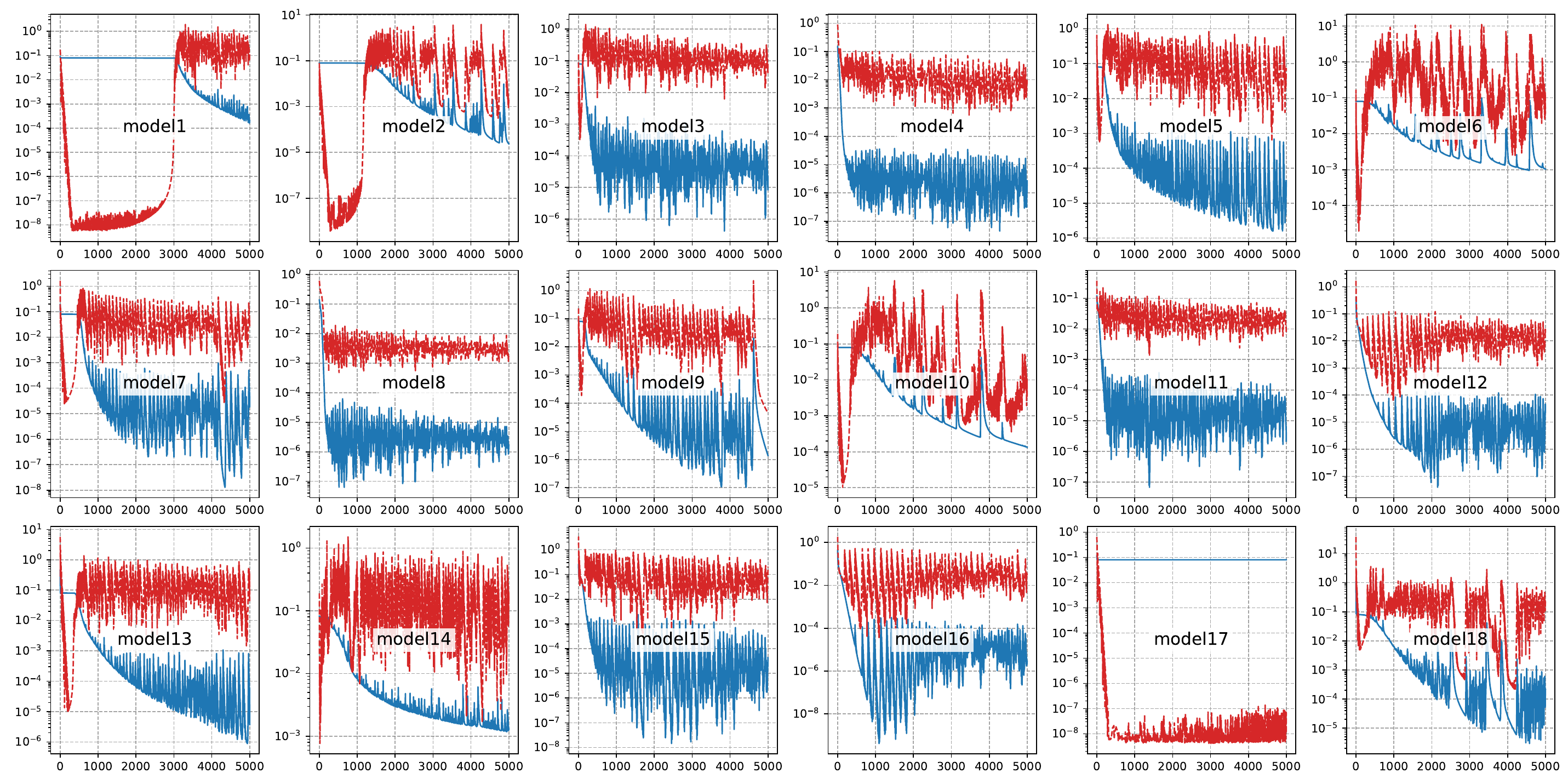}
  \centering
\caption{Training loss (solid blue) and L2 gradient norm (dashed red) during 5{,}000 training epochs for the 18 SechKAN variants. The variants differ in normalization placement (Norm1 or Norm2) and the use of a learnable basis width, as summarized in \Cref{tab:sechkan_variants}.}
\label{fig:grad_norm_loss}
\end{figure*}

\Cref{tab:sechkan_variants} summarizes the optimization behavior of the 18 SechKAN variants, while \Cref{fig:grad_norm_loss} shows the corresponding training loss and L2 gradient norm over 5,000 epochs. Without normalization (Models1--2), both models converge but exhibit different optimization trajectories. As shown in \Cref{fig:grad_norm_loss}, Model1 remains in a prolonged low-gradient regime for approximately the first 3,000 epochs, during which the training loss decreases only marginally. After this plateau, the gradient norm increases abruptly, allowing the loss to decrease rapidly before converging with mild oscillations. In contrast, the learnable basis width in Model2 shortens this low-gradient phase to approximately 1,000 epochs, enabling earlier loss reduction. Consequently, the best loss decreases from $1.61\times10^{-4}$ to $2.39\times10^{-5}$, while the final gradient norm is reduced by nearly two orders of magnitude. Except for Model17, all normalized variants largely eliminate this prolonged optimization plateau, with the loss decreasing substantially within the first 500 training epochs.

Norm1 (Models3--10) generally improves optimization accuracy and gradient stability. BatchNorm provides the most consistent optimization, producing the smallest gradient fluctuations and best losses below $10^{-7}$. LayerNorm and RMSNorm also converge reliably, whereas Min--Max normalization exhibits larger gradient variability and poorer optimization performance. Norm2 (Models11--18) is more sensitive to the normalization strategy. LayerNorm and BatchNorm remain effective, with Model16 achieving the lowest loss ($4.17\times10^{-10}$). In contrast, RMSNorm combined with the learnable basis width causes gradient collapse (Model17), while Min--Max normalization with the learnable basis width produces severe gradient spikes (Model18).

The optimization trajectories in \Cref{fig:grad_norm_loss} are consistent with these quantitative observations. Successful configurations maintain bounded gradient norms while steadily reducing the loss, whereas unstable configurations exhibit either repeated gradient explosions (Models6, 10, and 18) or premature gradient collapse (Model17). These results suggest that appropriate normalization mitigates activation saturation by maintaining informative gradients throughout training, although its effectiveness depends on both the normalization method and its placement. In general, BatchNorm provides the most stable gradient propagation, whereas Min--Max normalization is the most sensitive to normalization position and the learnable basis width.



\section*{Acknowledgements}

This research was supported by the VNUHCM-University of Information Technology's Scientific Research Support Fund.

\section*{Data Availability Statement}
\newcommand*{\brokenurlwithoutpar}[2]{{\texttt{#1}}\\*{\texttt{#2}}}
\begin{itemize}
    \item Our source code is publicly available at \url{https://github.com/hoangthangta/All-KAN}.
\end{itemize}

\section*{Compliance with Ethical Standards}
\begin{itemize}
    \item This article does not contain any studies with human participants or animals performed by any of the authors.
    \item All authors certify that they have no affiliations with or involvement in any organization or entity with any financial interest or non-financial interest in the subject matter or materials discussed in this manuscript.
\end{itemize}

\bibliographystyle{elsarticle-num-names}
\bibliography{elsarticle-template-num-names}

\end{document}